%% file: main.tex
\documentclass{article}
\usepackage{iclr2027_conference,times}
\setcitestyle{numbers,square,sort&compress}

\input{math_commands.tex}

\usepackage{amsmath}
\usepackage{amssymb}
\usepackage{graphicx}
\usepackage{float}
\usepackage{multirow}
\usepackage[hidelinks]{hyperref}
\usepackage{url}
\usepackage{booktabs,tabularx,array}
\usepackage{algorithm}
\usepackage{algpseudocode}

\usepackage{booktabs,tabularx,array}

\let\cite\citep

\newcommand{\method}{\textsc{PhysWAM}}

\title{\method{}: Physically Consistent \\ World Action Model for Autonomous Driving}

\author{%
  \rule{0pt}{34pt}\bfseries Dhruv Parikh$^{1,*}$ \quad Fengcheng Yu$^{1}$ \quad Quankai Gao$^{1}$ \quad Jiawei Yang$^{1}$ \quad Junjie Ye$^{1}$ \\[2pt]
  \bfseries Maulik Bhatt$^{2}$ \quad Thang Vu$^{2}$ \quad Charles Ochoa$^{3}$ \quad Rowan McAllister$^{3}$ \\[2pt]
  \bfseries Igor Vasiljevic$^{3}$ \quad Rajgopal Kannan$^{4}$ \quad Viktor Prasanna$^{1}$ \quad Vitor Guizilini$^{3}$ \quad Yue Wang$^{1}$ \\[5pt]
  \normalfont\small $^{1}$University of Southern California \quad $^{2}$Woven by Toyota \quad $^{3}$Toyota Research Institute \\[1pt]
  \normalfont\small $^{4}$DEVCOM Army Research Office \quad $^{*}$Corresponding author: \texttt{dhruvash@usc.edu}
}

\newif\ifpreprint
\preprinttrue
\ifpreprint
  \iclrfinalcopy
\fi


\selectfont   
\AtBeginDocument{\setlength{\parskip}{3pt plus 1pt minus 1pt}}
\usepackage[font=small,skip=2pt]{caption}
\makeatletter
\def\section{\@startsection {section}{1}{\z@}{-1.5ex plus -0.4ex minus -0.2ex}{1.0ex plus 0.2ex}{\large\bf\raggedright}}
\def\subsection{\@startsection{subsection}{2}{\z@}{-1.3ex plus -0.4ex minus -0.2ex}{0.8ex plus 0.2ex}{\normalsize\bf\raggedright}}
\makeatother
\begin{document}

\maketitle

\begin{figure}[H]
  \vspace{-2mm}
  \centering
  \setlength{\tabcolsep}{0pt}
  \renewcommand{\arraystretch}{0}
  \begin{tabular}{@{}c@{\hspace{0.02\linewidth}}c@{}}
    \includegraphics[width=0.48\linewidth]{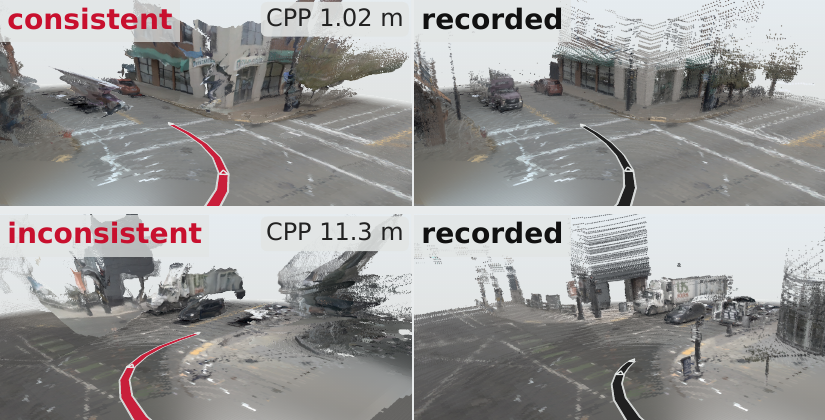} &
    \includegraphics[width=0.48\linewidth]{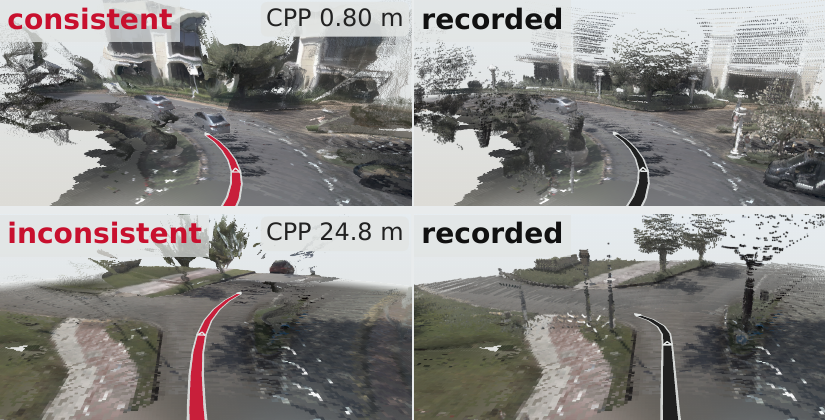} \\[0.012\linewidth]
    \includegraphics[width=0.48\linewidth]{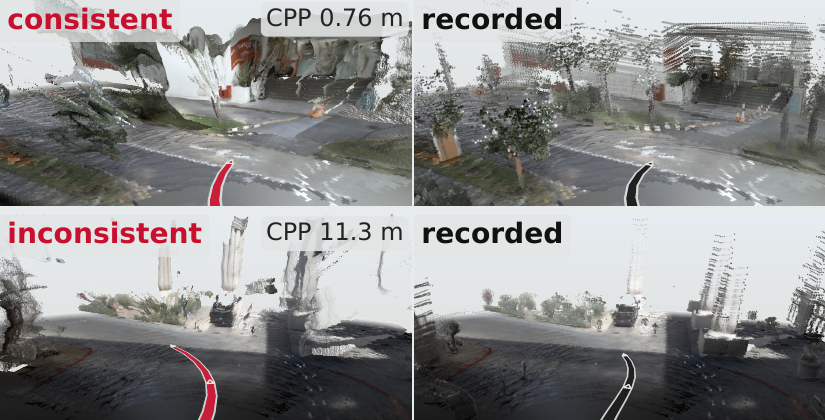} &
    \includegraphics[width=0.48\linewidth]{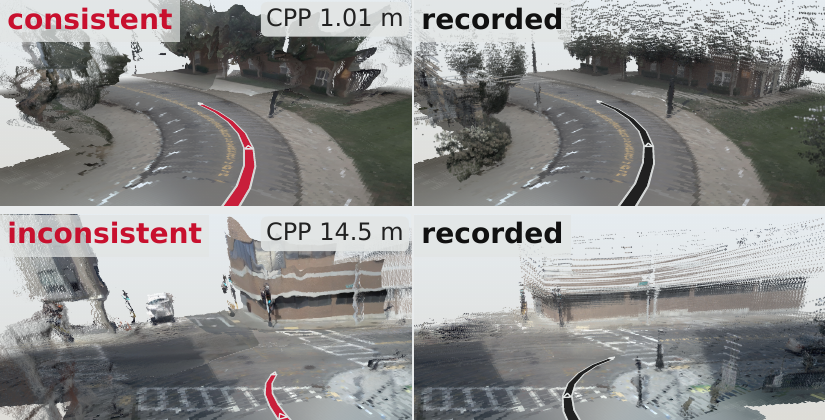}
  \end{tabular}
  \caption{
\textbf{Physical consistency through Coupled Point Projection (CPP).}
CPP jointly supervises generated depth and ego motion against measured scene geometry.
Generated depth is unprojected into 3D and transformed using generated motion (left).
LiDAR observations transformed using recorded motion provide the reference in the same coordinate frame (right).
The resulting geometric loss supervises both predictions: errors in depth, motion, or both can distort or displace the generated scene relative to the measured reference.
Red and black paths show generated and recorded motion, respectively.
First and third rows show close alignment with the measured scene; second and fourth rows show larger geometric discrepancies.
}
  \label{fig:teaser}
\end{figure}

\ifpreprint
  \lhead{Preprint}
\fi

\begin{abstract}
\input{sections/abstract}
\end{abstract}

\input{sections/introduction}
\input{sections/related_work}
\input{sections/method_v3}
\input{sections/experiments}

\input{sections/conclusion}

\input{sections/ai_use_statement}
\input{sections/statements}

\bibliography{references}
\bibliographystyle{plainnat}  

\clearpage
\appendix

\input{sections/appendix_method}
\input{sections/appendix_data}

\input{sections/appendix_experiments}
\input{sections/appendix_limitations}
\input{sections/appendix_related_work}

\end{document}

%% file: math_commands.tex
\usepackage{amsmath,amsfonts,bm}

\def\eqref#1{equation~\ref{#1}}

\def\1{\bm{1}}

\DeclareMathAlphabet{\mathsfit}{\encodingdefault}{\sfdefault}{m}{sl}
\SetMathAlphabet{\mathsfit}{bold}{\encodingdefault}{\sfdefault}{bx}{n}



%% file: sections/abstract.tex
World-action models (WAMs) jointly predict how a scene will evolve and how an agent should act, however joint generation alone does not necessarily impose a shared geometric constraint on these predictions.
We present \method{}, a unified world-action model for autonomous driving that co-denoises multiview video, metric depth, and ego motion within a single flow-matching transformer.
To ground world and action generation in measured scene geometry, we introduce Coupled Point Projection (CPP) that unprojects the generated depth into 3D points, transforms them using the generated $\mathrm{SE}(3)$ ego motion, and minimizes their distance to LiDAR points transformed using the recorded ego motion.
This geometric constraint promotes physical consistency with the measured scene by jointly supervising generated depth and motion alongside their standard flow-matching objectives.
At inference, trajectory selection relies only on a simple label-free consensus rule, with no learned scorer or simulator feedback.
We evaluate \method{} across NAVSIM v1 and v2 planning, zero-shot closed-loop transfer, and future video and metric-depth prediction.
Despite \method{}'s simple selection procedure, it achieves strong planning performance and transfers zero-shot to unseen driving environments.
It also generates accurate metric depth and temporally coherent video, with CPP improving both planning and depth prediction.
Together, these results demonstrate that the geometric relationship between scene depth and ego motion provides a direct way to couple world and action generation within a simple unified model.

%% file: sections/introduction.tex
\section{Introduction}
\label{sec:introduction}

A driving decision changes both how the ego vehicle moves and what it observes along its route.
A planned trajectory and the scene observed along it therefore describe the same future from different perspectives.
End-to-end driving systems learn such decisions directly from sensor observations~\cite{transfuser,uniad,vad,diffusiondrive}, driving world models generate future scenes~\cite{drivedreamer,gaia1,drivewm,vista}, increasingly across views and with geometric conditioning~\cite{magicdrive,omninwm}, and recent world-action models predict future observations and ego motion together~\cite{epona,driveva,drivewam}.
Joint generation alone, however, does not impose a shared geometric constraint on these predictions.

Existing approaches connect scene prediction and planning by using predicted images or features to shape a policy's representation~\cite{law,drivevla_w0,drivejepa,simwam}, by exposing present or predicted geometry to the planner~\cite{dvgt2,geoworldad,geowam}, by adding depth and other scene targets alongside video and action learning~\cite{drivedreamer_policy,explorevla,eponav2}, or by supervising the generated video with geometry~\cite{fourdwam}.
These approaches establish the value of scene prediction for driving.
The geometric relationship between scene depth and ego motion also lets us supervise the scene they describe together, using measured geometry as a common reference.
This motivates a shared objective that depends on both generated depth and motion, complementing their individual prediction losses.
Table~\ref{tab:paradigms} in App.~\ref{app:related_work} summarizes how representative works connect future prediction to planning.
This naturally motivates the question: \emph{how can we use the geometric relationship between scene depth and ego motion to couple world and action generation?}

Metric depth offers a direct way to express this relationship: feed-forward reconstruction methods recover scene structure and camera geometry from images~\cite{dust3r,vggt,mapanything}, and dynamic reconstruction methods recover geometry and motion over time from video~\cite{monst3r,d4rt}.
We draw on this geometric perspective to connect world and action generation.
Generated depth can be unprojected into 3D and transformed using generated ego motion into a shared reference frame.
LiDAR observations transformed using recorded motion provide the measured scene geometry in that same frame.
Errors in depth, motion, or both can distort or displace the generated scene relative to this reference.
We promote physical consistency with the measured scene by penalizing this discrepancy through a single geometric loss that supervises both predictions.
Figure~\ref{fig:teaser} visualizes this comparison for generated futures with small and large geometric discrepancies.

We introduce \method{}, a unified world-action model that co-denoises multiview video, metric depth, and ego motion within a single transformer.
Building on the Cosmos family of pretrained world models~\cite{cosmos,cosmos3}, we represent the three modalities in a shared sequence trained with flow matching~\cite{flowmatching}.
Conditioned on current multiview images, camera calibration, driving context, and ego history, the model jointly generates future video, metric depth maps, and an ego trajectory through iterative denoising.
Bidirectional attention allows all three modalities to exchange information throughout this process.

To explicitly couple the depth and ego motion generated by this model, we introduce \emph{Coupled Point Projection} (CPP).
CPP unprojects generated depth into 3D and transforms the resulting points using generated $\mathrm{SE}(3)$ ego motion.
Corresponding LiDAR observations transformed using recorded motion provide the reference in the same coordinate frame.
CPP penalizes distances between the generated and measured geometry, so errors in depth, motion, or both contribute to the same loss.
The supervision of each prediction therefore depends on the other, complementing their individual flow-matching objectives.
This promotes physical consistency with the measured scene through the geometric relationship between generated depth and ego motion. We also keep inference simple.
Several driving systems use learned scorers to evaluate candidate trajectories, either directly or from predicted future states~\cite{hydramdp,sparsedrivev2,wote,dawam}.
For each input scene, \method{} instead samples multiple trajectories and selects a representative one using only their pairwise distances.
This parameter-free consensus rule requires no labels, learned scorer, or simulator feedback.

We evaluate \method{} both as a driving policy and through the quality of its generated future world.
Planning experiments cover NAVSIM v1 and v2, including \texttt{navtest} and two-stage pseudo-simulation~\cite{navsim,pseudosim}.
Motivated by recent evidence of cross-domain driving transfer~\cite{driveva,beyonddrive}, we also evaluate zero-shot closed-loop driving in HUGSIM~\cite{hugsim}, without target-domain fine-tuning.
Beyond planning, we assess future-video quality and metric-depth accuracy.
Despite its simple trajectory selection, \method{} achieves strong planning compared to leading methods and transfers zero-shot across driving environments.
It also generates accurate metric depth and temporally coherent video.
CPP improves both planning and depth prediction, while consensus selection improves planning over using a single generated trajectory for each input scene.
Together, these results demonstrate the value of coupling generated depth and ego motion through a shared geometric objective. 

Our contributions are threefold.
\textbf{Unified world-action generation:} we develop a shared flow-matching formulation that co-denoises multiview video, metric depth, and ego motion within one model.
\textbf{Physical consistency through geometric coupling:} we introduce Coupled Point Projection (CPP), a geometric objective that jointly supervises generated depth and ego motion against measured scene geometry, complementing their individual prediction objectives.
\textbf{Broad evaluation with simple inference:} we demonstrate strong planning and zero-shot closed-loop transfer with label-free trajectory selection, alongside accurate metric depth and temporally coherent video.

%% file: sections/related_work.tex
\section{Related Work}
\label{sec:related_work}

\noindent\textbf{Generative driving world models.}
Driving world models generate controllable future video~\cite{drivedreamer,gaia1,vista}, across cameras in GAIA-2~\cite{gaia2}, and with LiDAR, occupancy, depth or semantics added to the generated future in MUVO, OccWorld and OmniNWM~\cite{muvo,occworld,omninwm}.
\method{} generates multiview video and metric depth together with ego motion in one flow-matching model.

\noindent\textbf{World-action models and planning.}
Future prediction improves policies as an auxiliary objective~\cite{law,simwam} and through joint video--action generation~\cite{epona,driveva,uwm,dreamzero}; predicted futures also serve learned trajectory evaluation, refinement and safety-aware exploration~\cite{world4drive,resim,driveworld_vla,explorevla}.
\method{} follows joint generation but selects among samples with a parameter-free consensus rule, without a learned scorer or simulator feedback.

\noindent\textbf{Geometry in world and action generation.}
Geometry enters generative driving models as a conditioning signal along supplied trajectories~\cite{geodrive,recamdriving}, as a generated output~\cite{unifuture,xwam,drivedreamer_policy}, as a planning representation~\cite{geoworldad,geowam}, or as a training target for the video features~\cite{geometryforcing,fourdwam,vggt,vggt-omega}, a line that goes back to learning depth and camera motion from view synthesis~\cite{sfmlearner}.
\method{} adds generated ego motion to the geometric objective: CPP unprojects the generated depth, carries it with the generated motion, and compares it with LiDAR under the recorded motion in one frame, supervising both outputs alongside their flow-matching objectives (App.~\ref{app:related_work} discusses these works in more detail).

%% file: sections/method_v3.tex
\section{Method}
\label{sec:method}

\begin{figure}[t]
    \centering
    \includegraphics[width=\linewidth]{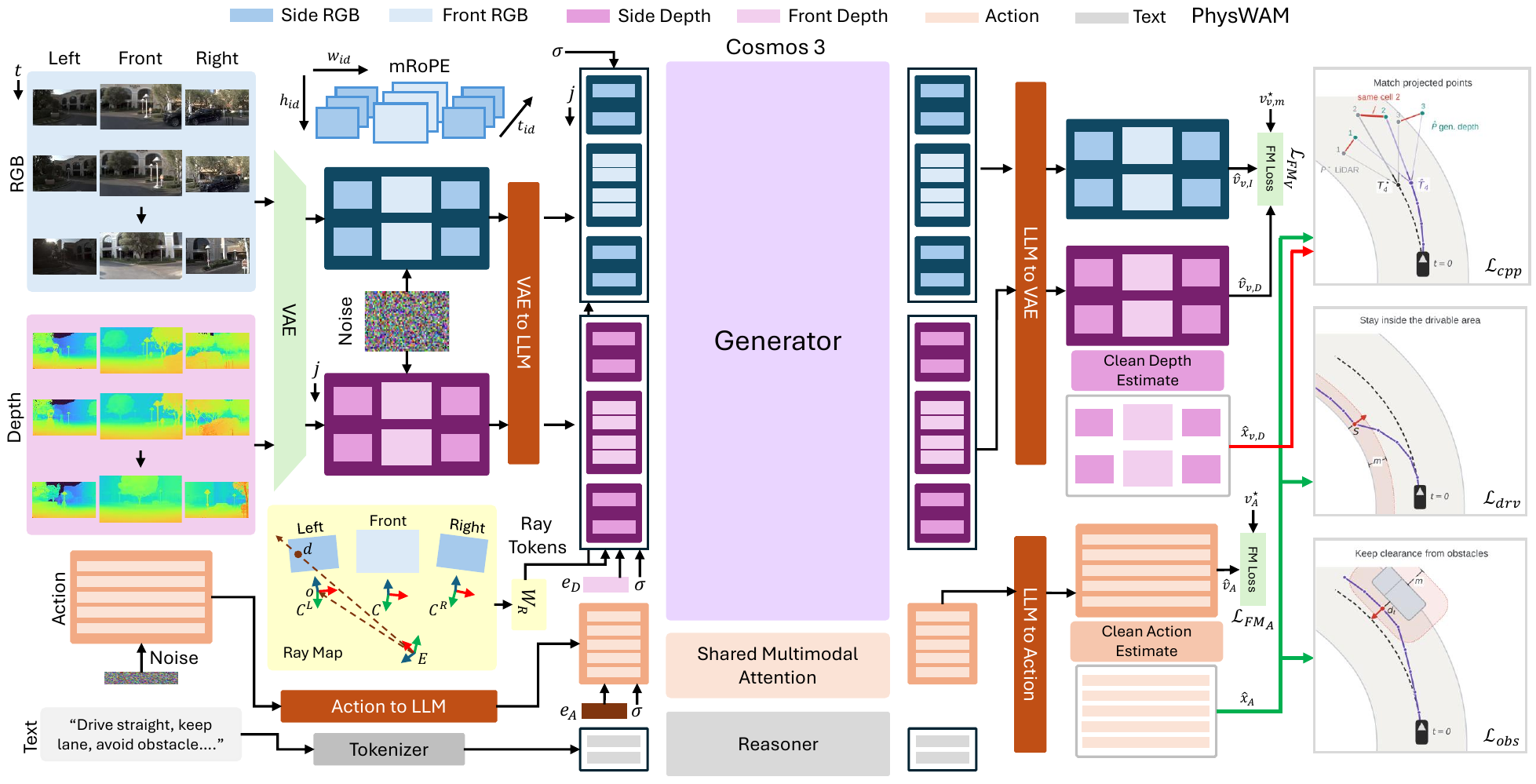}
    \caption{
        \textbf{Overview of \method{}.}
        The Cosmos~3 generator~\cite{cosmos3} jointly denoises multiview video, metric depth and ego motion, conditioned on text through its frozen reasoner; RGB and depth share a frozen VAE, and calibrated ray embeddings with mRoPE encode camera geometry and position.
        Current RGB and motion history stay clean; the rest is noised.
        Flow matching supervises the velocity outputs, whose clean estimates feed CPP, which transforms generated depth with generated motion and compares it with LiDAR under recorded motion, and the two hinges.
    }
    \label{fig:method_overview}
\end{figure}

\method{} combines joint video, depth, and motion generation with a geometric objective that supervises the depth--motion pair against measured scene structure (Fig.~\ref{fig:method_overview}).
We first define the generation problem and its representation, then introduce CPP, supporting trajectory constraints, and the training and inference procedures.

\subsection{Problem Formulation and Preliminaries}
\label{sec:method:preliminaries}

\noindent\textbf{World-action modeling with metric geometry.}
We study a driving world-action model that generates future observations and ego motion together with explicit scene geometry.
Let $\mathcal V=\{\mathrm{F},\mathrm{L},\mathrm{R}\}$ denote the front, left, and right cameras, and let $t=0,\ldots,T$ index scene frames, with $t=0$ the current time.
For view $v$, $\mathbf I_t^v$ is an RGB image and $\mathbf D_t^v$ is optical-axis depth in metres.
The context $c$ contains current images $\{\mathbf I_0^v\}_{v\in\mathcal V}$, camera calibration, textual driving context, and recent ego motion $\mathbf a_0$.
Starting from a pretrained world model~\cite{cosmos3}, we learn a conditional generative model $p_\theta(\{\mathbf I_{1:T}^v,\mathbf D_{0:T}^v\}_{v\in\mathcal V},\mathbf a_{1:T}\mid c)$, where $\mathbf a_t$ represents relative ego motion ending at time $t$.
Depth is generated at both current and future times; current RGB, motion history provide clean conditioning.

\noindent\textbf{Coordinate conventions.}
The front camera is the reference view: $\mathcal C_t$ is its frame, $\mathcal C_t^v$ the frame of view $v$, $\mathcal E_t$ the ego frame, and $\mathbf T_{A\leftarrow B}$ maps coordinates from $B$ to $A$.
The camera-to-ego calibration $\mathbf G_v=\mathbf T_{\mathcal E_t\leftarrow\mathcal C_t^v}$, fixed within a sequence, gives $\mathbf E_v=\mathbf G_{\mathrm F}^{-1}\mathbf G_v$, which maps view $v$ into the front camera at the same time. Each relative motion is $\boldsymbol\Delta_t=\mathbf T_{\mathcal C_{t-1}\leftarrow\mathcal C_t}$, and the composition $\mathbf T_t=\boldsymbol\Delta_1\cdots\boldsymbol\Delta_t$, with $\mathbf T_0=\mathbf I$, places the future front-camera frame in $\mathcal C_0$.

\noindent\textbf{Conditional flow matching.}
We parameterize $p_\theta$ with the generative transformer of a pretrained world model~\cite{cosmos3} and train it using rectified flow matching~\cite{rectifiedflow}.
Let $\mathbf x$ collect the continuous representations to be generated: VAE latents for RGB and depth, and pose coordinates for motion, defined in Sec.~\ref{sec:method:unified}.
At noise level $\sigma\in[0,1]$,
\begin{equation}
    \mathbf x^\sigma=(1-\sigma)\mathbf x+\sigma\boldsymbol\epsilon,
    \qquad
    \mathbf v^\star=\boldsymbol\epsilon-\mathbf x,
    \qquad
    \boldsymbol\epsilon\sim\mathcal N(\mathbf 0,\mathbf I).
    \label{eq:method:flow_path}
\end{equation}
One noise level is shared across the generated modalities of an example, with independently drawn Gaussian entries.
The transformer predicts the joint velocity $\widehat{\mathbf v}=v_\theta(\mathbf x^\sigma,\sigma;c)$.
These operations apply only to generated entries: conditioning frames and motion history are kept clean.
The estimated clean representation is
\begin{equation}
    \widehat{\mathbf x}
    =\mathbf x^\sigma-\sigma\widehat{\mathbf v}
    =\mathbf x+\sigma(\mathbf v^\star-\widehat{\mathbf v}).
    \label{eq:method:clean_estimate}
\end{equation}
This identity connects velocity prediction to the geometric objectives in Secs.~\ref{sec:method:cpp}--\ref{sec:method:hinges}: they act on clean depth and motion estimates from current forward pass, without complete denoising rollout.

\subsection{Unified World-Action--Geometry Denoising}
\label{sec:method:unified}

\noindent\textbf{Model architecture.}
We adapt the Cosmos~3 Mixture-of-Transformers architecture~\cite{cosmos3,mot}, based on Qwen3-VL~\cite{qwen3vl}.
Its understanding and generation pathways have separate parameters with shared attention: text attends causally within the text context, while visual and action tokens attend to text and one another bidirectionally.
We fine-tune the generation pathway, its projections, and embeddings, keeping the text pathway and video VAE fixed.
We extend this architecture to jointly denoise multiview video, depth, and motion without a separate depth denoiser.

\noindent\textbf{Visual and motion representations.}
The frozen causal video VAE~\cite{wan}, with encoder $\mathcal E$ and decoder $\mathcal D$, yields $\mathbf x_{v,I}=\mathcal E(\mathbf I_{0:T}^v)$ and $\mathbf x_{v,D}=\mathcal E(\mathcal C(\mathbf D_{0:T}^v))$ (depth labels: App.~\ref{app:data:depth}).
The fixed encoding $\mathcal C$ maps clipped log depth to the image range and repeats it across channels, preserving a common conversion to metres.
For both modalities, side-view height and width are half the front view's, reducing tokens while retaining lateral coverage.
Visual latents have shape $J\times h_v\times w_v\times c_\ell$: $j=0,\ldots,J-1$ indexes latent frames, $h_v,w_v$ are spatial dimensions, and $c_\ell$ is the channel width.
Temporal compression $s_t$ associates latent frame $j$ with scene frame $\kappa(j)=s_tj$, including $j=0$ for the current frame.
Actions retain the pretrained pose representation~\cite{cosmos3}, $\mathbf x_A=\mathbf a_{0:T}$ with $\mathbf a_t=[\mathbf t_t;\mathbf r_t^{(1)};\mathbf r_t^{(2)}]$ and each component in $\mathbb R^3$.
Translation is unnormalized in metres in the preceding front-camera frame; Gram--Schmidt orthonormalization of the first two rotation columns~\cite{rotation6d} gives the relative transform $\boldsymbol\Delta(\mathbf a_t)$.

\noindent\textbf{Tokenization and camera conditioning.}
A visual stream $i=(v,m)$ identifies view $v$ and modality $m\in\{I,D\}$; $A$ denotes the action stream.
The operator $\mathcal P$ patchifies (flattens) spatial $p\times p$ patches of VAE cells $u=(u_h,u_w)$ into $p^2c_\ell$-dimensional vectors, with each patch $q=(q_h,q_w)$ mapped to one token.
Noise and velocity targets remain in $\mathbf x_i$; projection yields $\mathbf z_i\in\mathbb R^{N_i\times d_h}$, containing $N_i$ tokens of hidden width $d_h$.
For patch $q$, its image-region centre $\boldsymbol\xi_v(q)$ is determined by the VAE stride and patch size.
Calibration gives its unit viewing direction $\mathbf d_v(q)$ and camera centre $\mathbf o_v$ in the instantaneous ego frame.
The Pl\"ucker ray $\mathbf r_v(q)=[\mathbf d_v(q);\mathbf o_v\times\mathbf d_v(q)]$ conditions projected patch:
\begin{equation}
    \mathbf z^i_{j,q}
    =\mathbf W_V[\mathcal P(\mathbf x_i^\sigma)]_{j,q}
     +\mathbf b_V+\mathbf W_R\mathbf r_v(q),
    \qquad i=(v,m).
    \label{eq:method:visual_embedding}
\end{equation}
The ray embedding is shared across RGB and depth and repeated over time; depth additionally receives a learned embedding $\mathbf e_D$.
Actions use $\mathbf z^A_t=\mathbf W_A\operatorname{pad}_A(\mathbf a_t^\sigma)+\mathbf b_A+\mathbf e_A$, reusing the pretrained pose projection and embedding~\cite{cosmos3}.
The operator $\operatorname{pad}_A$ zero-pads to the projection's input width; padding coordinates are excluded from losses.
All generated tokens receive the shared noise embedding $\mathbf e_\sigma(\sigma)$; clean context tokens do not.
The new $\mathbf W_R$ and $\mathbf e_D$ are zero-initialized to preserve pretrained content embeddings.

\noindent\textbf{Spatial and temporal alignment.}
Following the pretrained model~\cite{cosmos3}, mRoPE~\cite{qwen2vl} encodes time, height, and width in attention queries and keys.
For scene frame rate $f$, let $\tau(k)=\tau_0+\alpha k/f$, with common offset $\tau_0$ after text and pretrained time scale $\alpha$ in coordinate units per second.
We assign
\begin{equation}
    \boldsymbol\rho^{v,I}_{j,q}=\boldsymbol\rho^{v,D}_{j,q}
       =\big(\tau(\kappa(j)),\,q_h+b_v^h,\,q_w+b_v^w\big),
    \qquad
    \boldsymbol\rho^A_t=\big(\tau(t),\,0,\,0\big).
    \label{eq:method:positions}
\end{equation}
RGB and depth share coordinates; actions align with their endpoint time, with $\kappa$ accounting for visual temporal compression.
The offsets $(b_v^h,b_v^w)$ arrange left--front--right token grids panoramically; calibrated rays supply camera geometry.

\noindent\textbf{Velocity outputs and flow-matching objectives.}
From final hidden tokens $\mathbf z_i^{\mathrm{out}}$, output projections return velocities in the original representation spaces, $\widehat{\mathbf v}_i=\mathcal P^{-1}(\mathcal O_V(\mathbf z_i^{\mathrm{out}}))$ and $\widehat{\mathbf v}_A=\operatorname{unpad}_A(\mathcal O_A(\mathbf z_A^{\mathrm{out}}))$.
The visual projection returns $p^2c_\ell$ values per patch; unpatchification restores the VAE grid for computing the flow-matching loss as well as clean estimates (via Eq.~(\ref{eq:method:clean_estimate})).
Let $\operatorname{MSE}$ average squared velocity error within each stream, assigning zero error to conditioning entries.
With view--modality weights $\omega_{v,m}$,
\begin{equation}
    \mathcal L_{\mathrm{FM},V}
      =\mathbb E\!\left[\sum_{v,m}\omega_{v,m}
      \operatorname{MSE}(\widehat{\mathbf v}_{v,m},\mathbf v_{v,m}^{\star})\right],
    \qquad
    \mathcal L_{\mathrm{FM},A}
      =\mathbb E\!\left[\operatorname{MSE}(\widehat{\mathbf v}_A,\mathbf v_A^\star)\right].
    \label{eq:method:native_losses}
\end{equation}
Expectations average over examples, noise levels, and Gaussian noise.
These losses supervise each output individually; CPP explicitly supervises their geometric relationship.

\subsection{Coupled Point Projection}
\label{sec:method:cpp}

Depth and ego motion jointly determine how the generated scene is placed in 3D.
CPP uses this relationship to supervise both predictions against measured scene geometry: it transforms generated depth with generated motion, then compares the result with LiDAR transformed using recorded motion.
This gives a shared metric objective without sampling a full future during training.

\noindent\textbf{Efficient metric-depth estimation.}
Applying geometric supervision through the full video VAE decoder would require decoding dense depth at every training update.
Instead, a lightweight convolutional depth head $\Pi_\phi$ maps the estimated clean depth latents $\widehat{\mathbf x}_{v,D}$ directly to metric depth on the VAE grid.
It applies three spatial convolutions with GELU activations to each latent frame and produces one log-depth value per cell without changing spatial resolution.
Writing $\ell_{\min}=\log(1+d_{\min})$ and $\ell_{\max}=\log(1+d_{\max})$ for the fixed depth-encoding bounds, $\widehat d_j^v(u)=\exp(\operatorname{clip}([\Pi_\phi(\widehat{\mathbf x}_{v,D})]_j(u),\ell_{\min},\ell_{\max}))-1$, where $u$ indexes a VAE latent grid cell, not a transformer patch, and $\widehat d_j^v(u)$ is optical-axis depth in metres.
We fit the depth head $\Pi_\phi$ offline using masked smooth $L_1$ regression in log space to geometric-mean LiDAR depth pooled within each valid latent cell, using both clean depth latents and estimated-clean latents sampled from the training noise distribution.
Its parameters are then kept fixed during world model training, while CPP gradients pass through $\Pi_\phi$ to supervise the generated depth latents.
Clean action estimates $\widehat{\mathbf a}_t$ are converted to relative transforms and composed as $\widehat{\mathbf T}_t=\boldsymbol\Delta(\widehat{\mathbf a}_1)\cdots\boldsymbol\Delta(\widehat{\mathbf a}_t)$.

\noindent\textbf{Generated and measured geometry.}
Let $\mathbf b_v(u)$ be the unit camera-frame ray through the image location corresponding to latent cell $u$.
Calibrated unprojection is $\mathcal U_v(d,u)=d\,\mathbf b_v(u)/b_{v,z}(u)$: dividing by the ray's optical-axis component makes its $z$ coordinate one, so multiplication by optical-axis depth $d$ gives the camera-frame 3D location.
These rays are evaluated at VAE-cell centres; the ray embeddings in Sec.~\ref{sec:method:unified} use the centres of the larger token patches.
We evaluate CPP at future anchor frames $\mathcal A\subseteq\{\kappa(j):j>0\}$ and denote their latent indices by $j(t)=\kappa^{-1}(t)$. The $t=0$ (or $j=0$) frame is treated as the reference for this supervision. For each anchor, let $d_t^{L,v}(u)$ be the geometric mean of the LiDAR depths projected into cell $u$, and let $\mathbf T_t^\star$ be the recorded front-camera pose in $\mathcal C_0$.
Generated and measured depths use the same calibrated cell ray:
\begin{equation}
    \widehat{\mathbf P}_{v,t}(u)
       =\widehat{\mathbf T}_t\mathbf E_v\,
         \mathcal U_v\!\left(\widehat d_{j(t)}^v(u),u\right),
    \qquad
    \mathbf P^\star_{v,t}(u)
       =\mathbf T_t^\star\mathbf E_v\,
         \mathcal U_v\!\left(d_t^{L,v}(u),u\right).
    \label{eq:method:cpp_geometry}
\end{equation}
At each anchor, collecting these locations across LiDAR-supported cells and views forms two point clouds in $\mathcal C_0$: the scene implied by generated depth and motion, and its measured counterpart.
Corresponding cells identify the locations compared by CPP.
Both clouds describe the same future time, so expressing them in $\mathcal C_0$ does not require moving objects to remain static across the sequence.
Calibration, LiDAR depths, and recorded poses are fixed supervision (App.~\ref{app:data:lidar}).

\noindent\textbf{Physical consistency through geometric coupling.}
Let $\Omega$ contain valid $(v, t, u)$ correspondences with LiDAR support.
We use a robust metric loss
\begin{equation}
    \mathcal L_{\mathrm{cpp}}
      =\frac{1}{|\Omega|}\sum_{(v,t,u)\in\Omega}
         \rho_\delta\!\left(
         \big\|\widehat{\mathbf P}_{v,t}(u)
                    -\mathbf P^\star_{v,t}(u)\big\|_2\right),
    \label{eq:method:cpp_loss}
\end{equation}
where $\rho_\delta$ is Huber penalty~\cite{huber}, normalized by its transition distance: $\rho_\delta(e)=e^2/(2\delta)$ for $e\le\delta$ and $e-\delta/2$ otherwise.
This retains metric units while reducing influence of large discrepancies.
We average over valid correspondences across views and anchors, then over examples with LiDAR support. Errors in depth, motion, or both affect the same comparison.
Motion supervision depends on the generated geometry being transformed, while depth supervision depends on the generated transformation.
Gradients pass through the depth head and the composed action transforms to the corresponding clean estimates.
CPP therefore promotes physical consistency \emph{with the measured scene} by jointly supervising how generated depth and motion represent it.
The flow-matching objectives retain direct supervision of the individual predictions.

\subsection{Training and Inference}
\label{sec:method:training_inference}

\noindent\textbf{Geometric hinge losses.}
\label{sec:method:hinges}
CPP supervises the generated depth--motion pair; two complementary hinge losses supervise waypoint positions through the generated motion alone, independently of generated depth.
Both act on the generated front-camera position $\widehat{\mathbf p}_t=\pi_{xz}(\operatorname{trans}(\widehat{\mathbf T}_t))$ projected onto the ground plane of $\mathcal C_0$, against obstacle and drivable-area labels placed in the same frame (App.~\ref{app:data:hinges}).
The obstacle term $\mathcal L_{\mathrm{obs}}$ is a squared hinge on the clearance of $\widehat{\mathbf p}_t$ from static LiDAR structure and annotated object footprints at time $t$: it penalizes clearance below a margin $m_{\mathrm{obs}}$, capped at the clearance the recorded trajectory kept, so that the recorded path incurs no loss even in narrow corridors.
The drivable-area term $\mathcal L_{\mathrm{drv}}$ is a squared hinge on the signed distance of $\widehat{\mathbf p}_t$ inside the drivable region, computed from a distance-field raster by bilinear interpolation, below a margin $m_{\mathrm{drv}}$.
Both margins apply to positions rather than vehicle footprints, and the labels supply training supervision only; see App.~\ref{app:method:hinges} for details.

\noindent\textbf{Combined objective.}
All geometric losses act on the clean estimates in Eq.~(\ref{eq:method:clean_estimate}).
We downweight heavily noised examples with $w(\sigma)=(1-\sigma)^2$ and define noise-weighted losses $\overline{\mathcal L}_k=\mathbb E[w(\sigma)\mathcal L_k]$ for $k\in\{\mathrm{cpp},\mathrm{obs},\mathrm{drv}\}$.
The combined objective is
\begin{equation}
    \mathcal L_{\mathrm{total}}
       =\lambda_V\mathcal L_{\mathrm{FM},V}
        +\lambda_A\big(\mathcal L_{\mathrm{FM},A}
             +s_{\mathrm{cpp}}\overline{\mathcal L}_{\mathrm{cpp}}
             +s_{\mathrm{obs}}\overline{\mathcal L}_{\mathrm{obs}}
             +s_{\mathrm{drv}}\overline{\mathcal L}_{\mathrm{drv}}\big),
    \label{eq:method:total_objective}
\end{equation}
where $\lambda_V,\lambda_A>0$ weight visual and action supervision and the $s_k$ balance the geometric terms on the current update.

\noindent\textbf{Output-gradient balancing.}
We balance gradient magnitudes~\cite{gradnorm} rather than loss values.
On every update, $s_k$ is set so that the gradient of each active geometric term at the action velocity output has a fixed fraction $\eta_k$ of the norm of the action flow-matching gradient, and a second gain on CPP's backward depth path imposes the same fraction of the depth flow-matching gradient at the depth output, leaving forward values and the action path unchanged.
The coefficients are measured with local backward passes to the velocity outputs, held fixed during the single full backward pass, and set to zero for inactive terms (App.~\ref{app:method:balancing}).

\noindent\textbf{Joint sampling and trajectory selection.}
At inference, we jointly denoise Gaussian initializations from $\sigma=1$ to $0$ with current RGB, motion history, and text fixed; the VAE decoder returns RGB and metric depth, and composed motion, mapped to ego coordinates, gives the trajectory.
From $K$ samples we select the medoid trajectory, the one with the smallest total planar distance to others (App.~\ref{app:method:inference}).
This parameter-free rule uses no measured future, learned scorer, or simulator feedback.

%% file: sections/experiments.tex
\section{Experiments}
\label{sec:experiments}

\label{sec:exp:setup}

\noindent\textbf{Data and model.}
We train on the 103{,}281 windows of the NAVSIM \texttt{navtrain} split~\cite{navsim,openscene}: the current frame and the eight following frames at 2\,Hz from the front, front-left, and front-right cameras (App.~\ref{app:data}).
Depth, LiDAR, obstacle, and drivable-area labels are used in training only; at inference the model receives images, calibration, text, and motion history.
\method{} is initialized from Cosmos~3 Nano~\cite{cosmos3}; its generation pathway (7.0B of 15.2B parameters) is fine-tuned and the rest of the model is fixed.

\noindent\textbf{Training.}
We train for 30{,}000 updates with batches of 44 windows on NVIDIA RTX PRO 6000 Blackwell GPUs, using AdamW with a peak learning rate of $2\times10^{-5}$ decayed linearly to zero, and evaluate an exponential moving average of the weights.
In Eq.~(\ref{eq:method:total_objective}), $\lambda_V=10$, $\lambda_A=20$, and the target fraction $\eta_k$ of every geometric term is $0.2$.
All hyperparameters are listed in App.~\ref{app:experiments:hparams}.

\noindent\textbf{Benchmarks.}
On NAVSIM \texttt{navtest} (12{,}146 scenes) we report PDMS, the v1 score, and EPDMS, the v2 score, from the official devkit (sub-scores in App.~\ref{app:experiments:hparams}).
On \texttt{navhard}~\cite{pseudosim} we report the two-stage pseudo-simulation score: the plan is scored in 450 recorded scenes and then in 5{,}462 synthetic scenes with reactive traffic (App.~\ref{app:experiments:navhard}).
For zero-shot closed-loop driving we run all 436 HUGSIM episodes~\cite{hugsim} without any target-domain training, replanning every 0.5\,s, and report route completion (RC) and HD-Score, weighting the four difficulty levels by their number of episodes.

\noindent\textbf{Inference.}
We sample with UniPC~\cite{unipc} for 30 steps without guidance; each sample denoises RGB, depth, and ego motion jointly, and its eight waypoints form the plan.
A plan costs 9.4 GPU-seconds per scene at 30 steps; 8 steps cost 4.4 and score within a point (App.~\ref{app:experiments:selection}).
Unless marked, a row is one sample per scene; medoid rows select among $K{=}8$ samples by the rule of Sec.~\ref{sec:method:training_inference}.

\begin{table}[t]
\centering
\caption{\textbf{NAVSIM v1 and v2 on \texttt{navtest}.}
Numbers of other methods are as reported in their papers. R34: ResNet-34 backbone; +L: LiDAR input. Bold: best per column, oracle row excluded. \method{} rows use one sample per scene unless otherwise specified; the sample-to-sample sd is 0.24 PDMS and 0.30 EPDMS.}
\label{tab:navsim_v1}
\scriptsize
\setlength{\tabcolsep}{0.5pt}
\renewcommand{\arraystretch}{0.85}
\begin{tabular*}{\linewidth}{@{\extracolsep{\fill}}l l@{\hspace{3pt}} cccccc@{\hspace{3pt}} cccccccccc@{}}
\toprule
 & & \multicolumn{6}{c}{NAVSIM v1} & \multicolumn{10}{c}{NAVSIM v2} \\
\cmidrule(lr){3-8}\cmidrule(l){9-18}
Method & Input & NC & DAC & TTC & C & EP & PDMS & NC & DAC & DDC & TLC & EP & TTC & LK & HC & EC & EPDMS \\
\midrule
\textit{Human} & -- & 100 & 100 & 100 & 99.9 & 87.5 & 94.8 & \multicolumn{10}{c}{--} \\
\midrule
\multicolumn{18}{@{}l}{\textit{End-to-end planners}} \\
TransFuser~\cite{transfuser_pami} & 3$\times$Cam+L & 97.7 & 92.8 & 92.8 & 100.0 & 79.2 & 84.0 & 96.9 & 89.9 & 97.8 & 99.7 & 87.1 & 95.4 & 92.7 & 98.3 & 87.2 & 76.7 \\
LTF~\cite{transfuser_pami} & 3$\times$Cam & \multicolumn{6}{c}{--} & 97.6 & 91.9 & 99.2 & 99.8 & 87.6 & 97.2 & 96.6 & 98.3 & 86.3 & 83.6 \\
Hydra-MDP++ (R34)~\cite{hydramdppp} & 3$\times$Cam & 97.6 & 96.0 & 93.1 & 100 & 80.4 & 86.6 & 97.2 & 97.5 & 99.4 & 99.6 & 83.1 & 96.5 & 94.4 & 98.2 & 70.9 & 81.4 \\
ARTEMIS~\cite{artemis} & 3$\times$Cam+L & 98.3 & 95.1 & 94.3 & 100 & 81.4 & 87.0 & 98.3 & 95.1 & 98.6 & 99.8 & 81.5 & 97.4 & 96.5 & -- & \textbf{98.3} & 83.1 \\
DiffusionDrive~\cite{diffusiondrive} & 3$\times$Cam+L & 98.2 & 96.2 & 94.7 & 100.0 & 82.2 & 88.1 & 98.2 & 95.9 & 99.4 & 99.8 & 87.5 & 97.3 & 96.8 & 98.3 & 87.7 & 84.5 \\
WoTE~\cite{wote} & 3$\times$Cam+L & 98.5 & 96.8 & 94.9 & 99.9 & 81.9 & 88.3 & \multicolumn{10}{c}{--} \\
BeyondDrive~\cite{beyonddrive} & 3$\times$Cam & 98.4 & 97.9 & 95.0 & 100.0 & 83.7 & 89.7 & 98.4 & 97.9 & 99.5 & 99.8 & 87.8 & 98.0 & 97.3 & 98.3 & 88.5 & 90.1 \\
DriveSuprim (R34)~\cite{drivesuprim} & 3$\times$Cam & 97.8 & 97.3 & 93.6 & 100 & 86.7 & 89.9 & 97.5 & 96.5 & 99.4 & 99.6 & 88.4 & 96.6 & 95.5 & 98.3 & 77.0 & 83.1 \\
\midrule
\multicolumn{18}{@{}l}{\textit{Vision--language--action models}} \\
AutoVLA~\cite{autovla} & 3$\times$Cam & 98.4 & 95.6 & 98.0 & 99.9 & 81.9 & 89.1 & \multicolumn{10}{c}{--} \\
ReCogDrive~\cite{recogdrive} & 3$\times$Cam & 97.9 & 97.3 & 94.9 & 100 & 87.3 & 90.8 & \multicolumn{10}{c}{--} \\
\midrule
\multicolumn{18}{@{}l}{\textit{World models and world-action models}} \\
Epona~\cite{epona} & 1$\times$Cam & 97.9 & 95.1 & 93.8 & 99.9 & 80.4 & 86.2 & \multicolumn{10}{c}{--} \\
PWM~\cite{pwm} & 1$\times$Cam & 98.6 & 95.9 & 95.4 & 100.0 & 81.8 & 88.1 & \multicolumn{10}{c}{--} \\
DriveVLA-W0~\cite{drivevla_w0} & 1$\times$Cam & 98.7 & 96.2 & 95.5 & 100.0 & 82.2 & 88.4 & \multicolumn{10}{c}{--} \\
DVGT-2~\cite{dvgt2} & 8$\times$Cam & 97.8 & 97.2 & 93.9 & 100 & 83.4 & 88.6 & 97.8 & 97.2 & 99.6 & \textbf{99.9} & 88.4 & 97.3 & 98.1 & 98.2 & 83.2 & 88.9 \\
DriveDreamer-Policy~\cite{drivedreamer_policy} & 3$\times$Cam & 98.4 & 97.1 & 95.1 & 100.0 & 83.5 & 89.2 & 98.4 & 97.1 & 99.5 & \textbf{99.9} & 87.9 & 97.7 & 97.6 & 98.3 & 79.4 & 88.7 \\
DriveVLA-W0 (anchors)~\cite{drivevla_w0} & 1$\times$Cam & 98.7 & \textbf{99.1} & 95.3 & 99.3 & 83.3 & 90.2 & 98.5 & \textbf{99.1} & 98.0 & 99.7 & 86.4 & 98.1 & 93.2 & 97.9 & 58.9 & 86.1 \\
DVGT-2-NAVSIM~\cite{dvgt2} & 8$\times$Cam & 98.7 & 97.9 & 95.8 & 100 & 84.3 & 90.3 & 98.7 & 97.9 & \textbf{99.7} & \textbf{99.9} & 87.9 & 98.0 & \textbf{98.2} & 98.2 & 77.0 & 89.6 \\
GeoWAM~\cite{geowam} & 8$\times$Cam & \multicolumn{6}{c}{--} & 98.7 & 97.7 & \textbf{99.7} & \textbf{99.9} & 87.0 & 98.1 & 97.9 & 98.3 & 86.8 & 90.2 \\
\midrule
\method{} & 3$\times$Cam & \textbf{99.0} & 97.5 & 98.5 & 99.8 & 88.7 & 91.4 & \textbf{99.0} & 97.5 & 99.3 & 99.8 & 88.7 & 98.5 & 96.5 & 98.6 & 90.5 & 90.3 \\
\method{}, medoid-of-8 & 3$\times$Cam & \textbf{99.0} & 97.4 & \textbf{98.6} & 99.9 & \textbf{89.6} & \textbf{91.7} & \textbf{99.0} & 97.4 & 99.2 & \textbf{99.9} & \textbf{89.6} & \textbf{98.6} & 95.8 & \textbf{98.7} & 90.5 & \textbf{90.4} \\
\method{}, \textit{oracle-of-8} & 3$\times$Cam & \textit{99.7} & \textit{99.3} & \textit{99.2} & \textit{99.9} & \textit{91.9} & \textit{95.3} & \textit{99.7} & \textit{99.3} & \textit{99.0} & \textit{99.8} & \textit{91.9} & \textit{99.2} & \textit{95.8} & \textit{98.7} & \textit{90.0} & \textit{94.1} \\
\method{}, without CPP & 3$\times$Cam & 98.2 & 97.0 & 97.6 & 99.8 & 88.0 & 89.6 & 98.2 & 97.0 & 99.0 & 99.8 & 88.0 & 97.6 & 96.1 & 98.6 & 90.5 & 88.4 \\
\bottomrule
\end{tabular*}
\end{table}

\subsection{Planning}
\label{sec:exp:planning}

We compare against three families: end-to-end planners trained on NAVSIM, including the strongest recent one, BeyondDrive~\cite{beyonddrive}; vision--language--action models~\cite{autovla,recogdrive}; and world-action models that plan through a generative model of the scene~\cite{epona,pwm,drivevla_w0,dvgt2,drivedreamer_policy,geowam,eponav2,fourdwam,drivefuture}.
\method{} is trained by supervised fine-tuning with the CPP and hinge terms alone; several of the strongest compared methods add reinforcement learning after imitation~\cite{autovla,recogdrive,eponav2} or a trajectory scorer distilled from the simulator~\cite{hydramdp,drivesuprim}.

\noindent\textbf{NAVSIM \texttt{navtest}.}
With a single sample per scene, \method{} reaches 91.4 PDMS and 90.3 EPDMS (Table~\ref{tab:navsim_v1}), among the highest reported values for methods trained without a learned trajectory scorer or reinforcement-learning post-training; the best end-to-end planners (BeyondDrive, 89.7 and 90.1), vision--language--action models (ReCogDrive, 90.8) and world-action models (DVGT-2-NAVSIM, 90.3 PDMS; GeoWAM, 90.2 EPDMS) lie below, and planners that add a scorer or RL post-training reach up to 93.5 PDMS with a ViT-L backbone~\cite{drivesuprim}.
The margin comes from the safety terms: no-collision (99.0) and time-to-collision (98.5) are among the highest reported under both protocols, and ego progress (88.7) exceeds the compared methods by 1.4 under v1.
Selecting the medoid of eight samples adds 0.3 PDMS; the per-scene best of the eight reaches 95.3, so the samples contain better plans than label-free selection recovers.

\begin{table}[t]
\centering
\caption{\textbf{\texttt{navhard} and HUGSIM.} (a)~NAVSIM v2 on \texttt{navhard}, two-stage pseudo-simulation with reactive traffic; EPDMS as reported, with the sources of reprinted values and the stage scores in Table~\ref{tab:app:navhard_stages}. (b)~Zero-shot closed-loop driving on HUGSIM, as reported: route completion (RC) and HD-Score by difficulty; BeyondDrive's overall is the unweighted mean over the four levels. \method{} is trained on NAVSIM only, one sample per planner call. Bold: best.}
\label{tab:navhard}
\scriptsize
\renewcommand{\arraystretch}{0.85}
\begin{minipage}[c]{0.40\linewidth}
\centering
\textbf{(a)} NAVSIM v2 on \texttt{navhard}\\[2pt]
\setlength{\tabcolsep}{1pt}
\begin{tabular}{@{}l l c@{}}
\toprule
Method & Input & EPDMS \\
\midrule
LTF~\cite{transfuser_pami}                   & 3$\times$Cam & 25.1 \\
DiffusionDrive~\cite{diffusiondrive}    & 3$\times$Cam & 27.5 \\
DriveVLA-W0 (Flow)~\cite{drivevla_w0}   & 1$\times$Cam & 24.4 \\
DVGT-2~\cite{dvgt2}                     & 8$\times$Cam & 31.7 \\
DriveFuture~\cite{drivefuture}          & 6$\times$Cam & 34.6 \\
4D-WAM~\cite{fourdwam}                  & 1$\times$Cam & 35.9 \\
EponaV2~\cite{eponav2}                  & 1$\times$Cam & 36.1 \\
GeoWAM~\cite{geowam}                    & 8$\times$Cam & 36.6 \\
\midrule
\method{}                      & 3$\times$Cam & 38.1 \\
\method{}, medoid-of-8                  & 3$\times$Cam & \textbf{39.8} \\
\method{}, without CPP                  & 3$\times$Cam & 36.2 \\
\bottomrule
\end{tabular}
\end{minipage}\hfill
\begin{minipage}[c]{0.58\linewidth}
\centering
\textbf{(b)} Zero-shot closed-loop driving on HUGSIM\\[2pt]
\label{tab:hugsim}
\setlength{\tabcolsep}{0.5pt}
\begin{tabular*}{\linewidth}{@{\extracolsep{\fill}}l cc cc cc cc cc@{}}
\toprule
 & \multicolumn{2}{c}{Easy} & \multicolumn{2}{c}{Medium} & \multicolumn{2}{c}{Hard} & \multicolumn{2}{c}{Extreme} & \multicolumn{2}{c}{Overall} \\
\cmidrule(lr){2-3}\cmidrule(lr){4-5}\cmidrule(lr){6-7}\cmidrule(lr){8-9}\cmidrule(l){10-11}
Method & RC & HD & RC & HD & RC & HD & RC & HD & RC & HD \\
\midrule
UniAD~\cite{uniad}           & 58.6 & 48.7 & 41.2 & 29.5 & 40.4 & 27.3 & 26.0 & 14.3 & 40.6 & 28.9 \\
VAD~\cite{vad}               & 38.7 & 24.3 & 27.0 & 9.9  & 25.5 & 10.4 & 23.0 & 8.2  & 27.9 & 12.3 \\
LTF~\cite{transfuser_pami}        & 68.4 & 52.8 & 40.7 & 24.6 & 36.9 & 19.8 & 25.5 & 8.1  & 41.4 & 24.8 \\
BeyondDrive~\cite{beyonddrive} & 76.8 & 65.6 & 43.0 & 31.4 & 35.5 & 26.3 & 29.6 & 16.2 & 46.2 & 34.8 \\
\midrule
\textbf{\method{}}           & 93.5 & 86.9 & 47.4 & 30.1 & 38.6 & 25.2 & 26.1 & 13.4 & \textbf{48.9} & \textbf{35.5} \\
\method{}, without CPP       & 92.6 & 85.4 & 44.9 & 26.8 & 37.8 & 23.9 & 25.4 & 12.0 & 47.5 & 33.4 \\
\bottomrule
\end{tabular*}
\end{minipage}
\end{table}

\noindent\textbf{NAVSIM \texttt{navhard}.}
Under two-stage pseudo-simulation (Table~\ref{tab:navhard}a), \method{} reaches 38.1 EPDMS with a single sample and 39.8 with the medoid of eight, above the reported world-action models (GeoWAM, 36.6; EponaV2, 36.1; 4D-WAM, 35.9) and well above the end-to-end planners of Table~\ref{tab:navhard} (DiffusionDrive, 27.5; LTF, 25.1).
The advantage is largest in the second stage, under reactive traffic, where its no-collision, drivable-area and time-to-collision terms are the highest of the compared methods (Table~\ref{tab:app:navhard_stages}).

\noindent\textbf{Zero-shot closed loop.}
On HUGSIM (Table~\ref{tab:hugsim}b), without any training on the simulator's scenes, \method{} reaches RC 48.9 and HD-Score 35.5, above UniAD, VAD, LTF, and BeyondDrive (46.2 and 34.8; as the unweighted mean over the four levels, BeyondDrive's convention, \method{} reads 51.4 and 38.9).
The advantage is concentrated on the easy scenarios; on the medium, hard, and extreme scenarios the best reported HD-Scores remain above \method{}'s (Table~\ref{tab:hugsim}b).
Drivable-area compliance is above 93.8 at every difficulty, and the score is limited by the no-collision and time-to-collision terms (Table~\ref{tab:app:hugsim_tiers}).

\begin{table}[t]
\centering
\caption{\textbf{Ablations and future depth.} (a)~The full recipe with and without CPP after 4{,}000 updates on \texttt{navtest} (off-road, collision: share of scenes with a zero drivable-area or no-collision term; heading: median error against the recorded trajectory) and at the end of training (Tables~\ref{tab:navsim_v1}--\ref{tab:hugsim}). (b,~d)~LoRA setting on \texttt{navtest}, one sample per scene: objective terms added to flow matching, and the front view alone against all three views by driving command, with the full model of Table~\ref{tab:navsim_v1} for reference. (c)~Generated depth of the full model against the LiDAR of the same future frame and camera, without scale alignment, four samples per scene; side views in Table~\ref{tab:app:cpp_depth}; lower block as reported on nuScenes by GeoWAM~\cite{geowam}.}
\label{tab:cpp}
\scriptsize
\renewcommand{\arraystretch}{0.85}
\begin{minipage}[c]{0.55\linewidth}
\centering
\textbf{(a)} CPP at the full recipe\\[2pt]
\setlength{\tabcolsep}{1pt}
\begin{tabular*}{\linewidth}{@{\extracolsep{\fill}}l ccccc@{}}
\toprule
 & PDMS & EPDMS & Off-road & Collision & Heading \\
\midrule
4k, with CPP    & 80.3 & 78.8 & 9.7\%  & 3.8\% & 2.7$^\circ$ \\
4k, without CPP & 57.0 & 55.7 & 32.7\% & 9.8\% & 8.9$^\circ$ \\
\midrule
 & \multicolumn{2}{c}{\texttt{navtest} PDMS\,/\,EPDMS} & \texttt{navhard} & \multicolumn{2}{c}{HUGSIM RC\,/\,HD} \\
\cmidrule(lr){2-3}\cmidrule(lr){4-4}\cmidrule(l){5-6}
final, with CPP    & \multicolumn{2}{c}{91.4 / 90.3} & 38.1 & \multicolumn{2}{c}{48.9 / 35.5} \\
final, without CPP & \multicolumn{2}{c}{89.6 / 88.4} & 36.2 & \multicolumn{2}{c}{47.5 / 33.4} \\
\bottomrule
\end{tabular*}\\[8pt]
\textbf{(b)} Objective terms\\[2pt]
\label{tab:lora}
\setlength{\tabcolsep}{3pt}
\begin{tabular*}{\linewidth}{@{\extracolsep{\fill}}l cc@{}}
\toprule
Objective & PDMS & EPDMS \\
\midrule
flow                    & 85.7 & 84.3 \\
flow + hinge            & 87.2 & 85.7 \\
flow + CPP              & 87.8 & 86.2 \\
flow + hinge + CPP      & \textbf{88.2} & \textbf{86.5} \\
\bottomrule
\end{tabular*}
\end{minipage}\hfill
\begin{minipage}[c]{0.43\linewidth}
\centering
\textbf{(c)} Future metric depth against LiDAR\\[2pt]
\label{tab:worldmodel}
\setlength{\tabcolsep}{2pt}
\begin{tabular*}{\linewidth}{@{\extracolsep{\fill}}l cc cc@{}}
\toprule
 & \multicolumn{2}{c}{+2\,s} & \multicolumn{2}{c}{+4\,s} \\
\cmidrule(lr){2-3}\cmidrule(l){4-5}
View & AbsRel$\downarrow$ & $\delta_{1.25}\uparrow$ & AbsRel$\downarrow$ & $\delta_{1.25}\uparrow$ \\
\midrule
Front        & 0.175 & 0.814 & 0.232 & 0.742 \\
Front, cells & 0.144 & 0.856 & 0.194 & 0.795 \\
\midrule
\multicolumn{5}{@{}l}{\textit{Reported on nuScenes, front view}} \\
Epona + DVGT    & 0.263 & 0.677 & 0.310 & 0.589 \\
Cosmos~3 + DVGT & 0.376 & 0.513 & 0.422 & 0.447 \\
VGGT-World      & 0.329 & 0.553 & 0.357 & 0.497 \\
GeoWAM          & 0.245 & 0.769 & 0.297 & 0.703 \\
\bottomrule
\end{tabular*}\\[8pt]
\textbf{(d)} Views (PDMS)\\[2pt]
\setlength{\tabcolsep}{2.5pt}
\begin{tabular*}{\linewidth}{@{\extracolsep{\fill}}l cccc@{}}
\toprule
Views & All & Left & Straight & Right \\
\midrule
single-view             & 87.1 & 81.2 & 90.0 & 82.0 \\
multi-view              & \textbf{88.2} & \textbf{83.9} & \textbf{90.3} & \textbf{84.3} \\
\midrule
full fine-tune & 91.4 & 87.8 & 92.7 & 90.4 \\
\bottomrule
\end{tabular*}
\end{minipage}
\end{table}

\begin{figure}[t]
\centering
\includegraphics[width=0.9\linewidth]{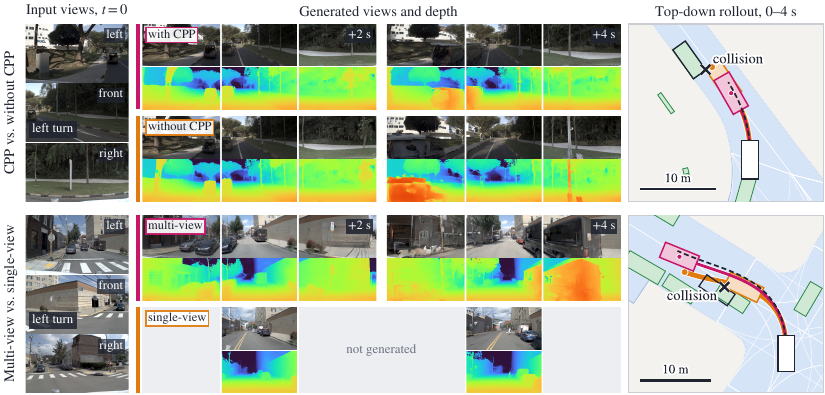}
\caption{\textbf{Qualitative examples.} Left: input views at $t=0$ and the command. Middle: generated views and metric depth at $+2$ and $+4$\,s; the single-view model generates the front view only. Right: rollouts over the drivable area (blue), recorded traffic at 4\,s (green), recorded trajectory (dashed); ego boxes at the end point or at the failure step, with the hit agent outlined. Top: with CPP the model generates a parked van a few metres off in its left view and stops short of it; without CPP it generates the van pressed against the ego and hits it at 3.4\,s. Bottom: an oncoming car visible only in the left camera; the multi-view model generates it passing close and keeps the turn wide, the single-view model cuts the corner and collides at 2.3\,s.}
\label{fig:qualitative}
\end{figure}

\subsection{Ablations}
\label{sec:exp:ablations}

We study the design of \method{} in two settings.
CPP is ablated at the full recipe above by training the same model without the CPP term, reported as ``without CPP'' in Tables~\ref{tab:navsim_v1}--\ref{tab:hugsim}.
The objective terms and the views are ablated in a lighter setting: rank-128 LoRA adapters~\cite{lora} on the generation pathway, trained for 16{,}000 updates, with all other differences from the full recipe listed in Table~\ref{tab:app:hparams_lora}.
Each row of this setting differs from the full objective only in terms or views it names.

\noindent\textbf{Coupled point projection.}
CPP speeds up learning to plan: after 4{,}000 updates, the model without CPP scores 57.0 PDMS against 80.3, leaves the drivable area in a third of the scenes rather than a tenth, and has more than three times the heading error (Table~\ref{tab:cpp}a, top).
The gap narrows with training but does not close: the final model without CPP scores 89.6 against 91.4 PDMS on \texttt{navtest}, 36.2 against 38.1 EPDMS on \texttt{navhard}, and 33.4 against 35.5 HD-Score on HUGSIM (Table~\ref{tab:cpp}a, bottom), below the best prior method on each (ReCogDrive, 90.8; GeoWAM, 36.6; BeyondDrive, 34.8).
It also generates worse depth and motion: its per-pixel depth error against LiDAR is 8--15\% higher across views and horizons (AbsRel 0.193 against 0.175 in the front view at +2\,s), and fewer of its sampled futures end in a pose that agrees with the recorded one (70.9\% against 81.4\%; Table~\ref{tab:app:cpp_depth}).

\noindent\textbf{Objective terms.}
Under the LoRA setting, the hinges add 1.5 PDMS to flow matching, CPP adds 2.1, and both together add 2.5 (Table~\ref{tab:lora}b).
CPP is the larger contribution, and the hinges still add 0.4 on top of it.
Flow matching supervises each modality in its native representation and the hinges supervise waypoint feasibility; CPP adds a shared metric-space residual whose gradients depend jointly on the generated depth and motion, so each branch is trained to agree with the other's current prediction as well as with the measurement (App.~\ref{app:method:cpp}).
The ablation establishes the benefit of this added objective; distinguishing the effect of the coupling from that of metric-space supervision of each branch requires a matched decoupled comparison.

\noindent\textbf{Multi-view generation.}
Observing and generating all three views instead of the front view alone raises PDMS by 1.1 overall (Table~\ref{tab:lora}d).
The gain is 0.3 on straight roads and 2.3--2.7 on turns, where the drivable boundary and crossing traffic lie in the side views that the single-view model neither observes nor generates. Without generated depth, and hence without CPP on the side-view cells, the three views add only 0.4 over the front view alone (App.~\ref{app:experiments:ablations}), so most of the gain arrives via CPP.

\noindent\textbf{Hinge schedule.}
Both hinges are one-sided and vanish on every plan that keeps its margins, so late in training the balancing of Sec.~\ref{sec:method:training_inference} would concentrate their fixed gradient fraction on few plans that still violate them; we therefore anneal both hinges to zero by the middle of training.
CPP penalizes deviation in either direction on every LiDAR-supported cell, so it does not vanish once plans are safe, and removing it from update 20{,}000 onward costs 1.2 HD-Score on HUGSIM (App.~\ref{app:experiments:ablations}).

\noindent\textbf{Qualitative examples.}
Figure~\ref{fig:qualitative} shows both ablations on individual scenes: without CPP the plan drives into a parked van that the full model stops short of, and without the side views the plan cuts a left turn into an oncoming car that only the left camera shows at $t=0$.


\subsection{World-Model Quality}
\label{sec:exp:world}

\noindent\textbf{Metric depth.}
Scored against the LiDAR sweep of the same future frame and camera without scale alignment, front-view depth reaches AbsRel 0.175 at +2\,s and 0.232 at +4\,s (Table~\ref{tab:worldmodel}c), and 0.144 and 0.194 on the cells that CPP supervises.
These values are below the future-depth results reported on nuScenes, a different dataset and rig (GeoWAM~\cite{geowam}, 0.245 and 0.297; Epona~\cite{epona} with a DVGT~\cite{dvgt} head, 0.263 and 0.310; VGGT-World~\cite{vggtworld}, 0.329 and 0.357; Cosmos~3~\cite{cosmos3} + DVGT, 0.376 and 0.422); the side views and the error structure are in App.~\ref{app:experiments:consistency}.

\noindent\textbf{Video.}
The generated front video reaches FVD-9 111.3 against a recorded-versus-recorded floor of 91.5 at 600 clips per side, and 42.2 over all 1{,}200 clips (Table~\ref{tab:app:video}); reported NAVSIM values range from 142.6 (DrivingGPT~\cite{drivinggpt}) to 32.7 (CoWorld-VLA~\cite{coworldvla}) under their own protocols.
The generated video and depth also agree with the generated motion: the camera yaw recovered from the video deviates from it by a median of $0.80^\circ$ over 4\,s against a $0.29^\circ$ floor on recorded clips, and the front and side depths agree to within 1.5$\times$ the LiDAR floor (App.~\ref{app:experiments:consistency}).

%% file: sections/conclusion.tex
\section{Conclusion}
\label{sec:conclusion}
We presented \method{}, a world-action model that denoises multi-view video, metric depth and ego motion in one generator and couples the generated depth and motion through Coupled Point Projection (CPP), one loss on projected LiDAR geometry in metric space, alongside two geometric hinge losses and output-gradient balancing.
The model plans strongly on NAVSIM, transfers zero-shot to closed-loop driving and generates accurate future depth. Removing CPP degrades both planning and depth performance.
A self-supervised form of the coupling would let the model scale to driving data without LiDAR or dense labels, and consistency on rollouts away from the strictly recorded trajectory. Additionally, backbones and scales beyond the one studied here, are the relevant directions we find promising to explore (limitations in App.~\ref{app:limitations}).

%% file: sections/ai_use_statement.tex
\subsection*{AI Use Statement}
AI models, including Claude and ChatGPT, were used as general-purpose assistants under the authors' supervision: for writing and polishing the paper, for implementing the method, and for running and monitoring the experiments.
The authors hold responsibility for the content of the paper.

%% file: sections/statements.tex
\subsection*{Ethics Statement}
This work uses publicly available driving datasets and pretrained models under their respective licenses and involves no human subjects and no data collection or annotation beyond what those datasets provide.
\method{} is a research model evaluated on recorded data and in simulation; it is not validated for deployment on public roads, and its trajectories are not meant to control a vehicle without the safety systems such deployment requires.

\subsection*{Reproducibility Statement}
\method{} builds on open-source datasets (NAVSIM and its OpenScene source data, HUGSIM for closed-loop evaluation) and on the publicly released Cosmos~3 Nano checkpoint, which we fine-tune.
We will release our code, training and evaluation scripts, and model checkpoints.
The appendices give the full details needed to reproduce the results: the architecture and objective (App.~\ref{app:method}), data and label construction (App.~\ref{app:data}), and hyperparameters and evaluation protocols (App.~\ref{app:experiments}).

%% file: sections/appendix_method.tex

\section{Additional Method Details}
\label{app:method}

This appendix follows the order of Sec.~\ref{sec:method}: the representations on which the losses act (Sec.~\ref{app:method:architecture}), the depth head and the gradients of CPP (Sec.~\ref{app:method:cpp}), the distance functions of the hinge losses (Sec.~\ref{app:method:hinges}), output-gradient balancing and the complete update (Sec.~\ref{app:method:balancing}), and inference (Sec.~\ref{app:method:inference}).
Notation follows the main text: $t$ indexes scene frames, $j$ latent frames, and $\sigma$ the noise level; $\mathbf x$ denotes VAE-grid or pose representations and $\mathbf z$ transformer tokens.

\subsection{Architecture and Representations}
\label{app:method:architecture}

\noindent\textbf{Pretrained architecture and adaptation.}
\method{} builds on the Cosmos~3 Mixture-of-Transformers architecture~\cite{cosmos3,mot}, based on Qwen3-VL~\cite{qwen3vl}.
Within each layer, understanding (text) tokens and generation tokens have separate normalization, query/key/value projections, output projections, and feed-forward parameters, and share one attention operation.
Text queries attend causally to text; generation queries attend bidirectionally to all text, visual, and action tokens of the same example.
Text and generated content therefore interact in every layer rather than through a completed prediction passed from one model to another.
All RGB and depth tokens, including the clean RGB context, use the generation pathway.
We adapt the generation pathway with its input projections, output projections, and embeddings, and keep the text pathway, the Wan video VAE~\cite{wan}, and the depth head of Sec.~\ref{app:method:cpp} fixed.
RGB and depth share the VAE and the visual projections; the depth embedding $\mathbf e_D$ distinguishes the two modalities and the ray embedding identifies their viewing geometry.

\noindent\textbf{Token layout.}
Table~\ref{tab:app:representations} follows a visual stream $i=(v,m)$ from image to velocity.
With $\bar h_v=\lceil h_v/p\rceil$ and $\bar w_v=\lceil w_v/p\rceil$, the stream has $N_i=J\bar h_v\bar w_v$ tokens, of which $N_i^{\mathrm{gen}}$ lie at generated frames.
Padding introduced for the VAE stride and for patchification is removed before any loss or depth estimate is formed, so all quantities below refer to the content grid.

\begin{table}[!ht]
    \centering
    \caption{\textbf{Visual representation pathway} for one stream $i=(v,m)$. Shapes omit the minibatch dimension. Noise and flow targets are defined on the VAE grid, before patchification; each spatial patch becomes one token.}
    \label{tab:app:representations}
    \small
    \setlength{\tabcolsep}{4pt}
    \renewcommand{\arraystretch}{1.08}
    \begin{tabularx}{\linewidth}{@{}l l >{\raggedright\arraybackslash}X@{}}
        \toprule
        \textbf{Quantity} & \textbf{Shape} & \textbf{Operation} \\
        \midrule
        RGB or encoded depth & $(T+1)\times H_v\times W_v\times 3$ & Input to the shared frozen VAE. \\
        $\mathbf x_i$ & $J\times h_v\times w_v\times c_\ell$ & VAE grid, indexed spatially by $u$. \\
        $\mathcal P(\mathbf x_i^\sigma)$ & $N_i\times p^2c_\ell$ & Flattened patches of the noisy grid, indexed by $q$. \\
        $\mathbf z_i$ & $N_i\times d_h$ & Projected patches with ray, modality, and noise embeddings. \\
        $\mathcal O_V(\mathbf z_i^{\mathrm{out}})$ & $N_i^{\mathrm{gen}}\times p^2c_\ell$ & Patch velocities at generated positions. \\
        $\widehat{\mathbf v}_i$ & $J\times h_v\times w_v\times c_\ell$ & Unpatchified velocity field; conditioning frames are zero-filled and not predicted. \\
        \bottomrule
    \end{tabularx}
\end{table}

\noindent\textbf{Metric depth encoding.}
The VAE receives depth as an image.
For fixed bounds $d_{\min},d_{\max}$, with $\ell_{\min}=\log(1+d_{\min})$ and $\ell_{\max}=\log(1+d_{\max})$, the scalar encoding is
\begin{equation}
    \mathcal C(d)=
    2\frac{\ell_{\max}-\log(1+\operatorname{clip}(d,d_{\min},d_{\max}))}
            {\ell_{\max}-\ell_{\min}}-1,
    \label{eq:app:depth_codec}
\end{equation}
so that nearer surfaces are brighter and the full range maps onto $[-1,1]$.
The scalar is quantized to the image format, replicated across the three channels, and normalized to the VAE input range.
Decoding averages the reconstructed channels and inverts the continuous map,
\begin{equation}
    \mathcal C^{-1}(y)=
    \exp\!\left(\ell_{\max}
      -\frac{\operatorname{clip}(y,-1,1)+1}{2}
       (\ell_{\max}-\ell_{\min})\right)-1.
    \label{eq:app:depth_inverse}
\end{equation}
Because the bounds are shared by all scenes, a decoded depth is metric without a per-scene scale; quantization and VAE reconstruction add approximation error but leave this convention intact.

\noindent\textbf{Camera geometry at two resolutions.}
With VAE spatial stride $s_x$, the image-plane centres of latent cell $u=(u_h,u_w)$ and of token patch $q=(q_h,q_w)$ are
\begin{equation}
    \boldsymbol\xi_v^{\mathrm{cell}}(u)
       =s_x\big(u_w+\tfrac12,\,u_h+\tfrac12\big),
    \qquad
    \boldsymbol\xi_v(q)
       =s_xp\big(q_w+\tfrac12,\,q_h+\tfrac12\big),
    \label{eq:app:ray_centres}
\end{equation}
in the coordinates of the resized image, with calibration scaled accordingly; the rays account for lens distortion.
For the ray embedding, the unit camera-frame ray through $\boldsymbol\xi_v(q)$ is rotated by $\operatorname{rot}(\mathbf G_v)$ into the ego frame and paired with the camera centre $\operatorname{trans}(\mathbf G_v)$ to form the Pl\"ucker coordinates $\mathbf r_v(q)$ of Eq.~(\ref{eq:method:visual_embedding}).
CPP uses the camera-frame ray $\mathbf b_v(u)$ through $\boldsymbol\xi_v^{\mathrm{cell}}(u)$.
The embedding and the unprojection thus share one camera model and differ only in resolution.

\noindent\textbf{Temporal coordinates.}
mRoPE~\cite{qwen2vl,cosmos3} assigns every token a (time, height, width) position.
The spatial offsets $(b_v^h,b_v^w)$ in Eq.~(\ref{eq:method:positions}) are in token-grid units and place the left, front, and right grids side by side; camera geometry enters only through the ray embedding.
Along time, successive visual latents advance by $\alpha s_t/f$ and successive actions by $\alpha/f$, so action $t=\kappa(j)$ and latent frame $j$ share a temporal coordinate although their sequence positions differ.
Every visual stream starts at the common offset $\tau_0$ after the text, and text tokens use their causal index on all three axes.
Scene time is thereby encoded in the attention geometry, while the noise level of generated tokens is carried separately by $\mathbf e_\sigma$.

\noindent\textbf{Stream losses.}
The current RGB latent and the motion history are clean; all depth latents and all future RGB and motion entries are generated, with one $\sigma$ per example and independent Gaussian entries across its streams.
For stream $i$, let $\mathcal I_i$ be the scalar entries of its retained (unpadded) representation and $M_i(e)\in\{0,1\}$ indicate a generated entry.
The stream loss in Eq.~(\ref{eq:method:native_losses}) is
\begin{equation}
    \mathcal L_{\mathrm{FM}}^{(i)}
       =\frac{1}{|\mathcal I_i|}
        \sum_{e\in\mathcal I_i}M_i(e)
        \big(\widehat{\mathbf v}_i(e)-\mathbf v_i^\star(e)\big)^2.
    \label{eq:app:fm_reduction}
\end{equation}
Conditioning entries contribute zero error but remain in the denominator, so the normalization of a stream does not depend on how many of its entries are generated.
For the action stream, $\mathcal I_A$ comprises the history row and the future rows over the nine pose coordinates.
Each visual stream is reduced to its mean before the weights $\omega_{v,m}$ are applied, so a front view with many tokens and a side view with few enter the visual loss on the same footing; the expectation in Eq.~(\ref{eq:method:native_losses}) then averages over examples.

\subsection{Depth Head and Coupled Point Projection}
\label{app:method:cpp}

\noindent\textbf{Depth head.}
The head $\Pi_\phi$ of Sec.~\ref{sec:method:cpp} stacks three $3\times3$ convolutions with channel widths $c_\ell\to c_\Pi\to c_\Pi\to1$ and GELU after the first two, at stride one with symmetric padding, so the VAE grid is preserved.
It acts on each latent frame independently, is shared by the three views, and takes the estimated clean latent $\widehat{\mathbf x}_{v,D}$ after unpatchification, in the normalization of the generative model.

We fit $\phi$ once, before world-model training, on clean latents $\mathbf x_{v,D}$ and on estimated-clean latents $\widehat{\mathbf x}_{v,D}$ computed from the velocity predictions of a preliminary model with the same architecture and representation, at noise levels drawn from the training distribution and weighted by $w(\sigma)$, so that the head is most accurate where CPP carries most weight.
The target of a latent cell with LiDAR depths $\{d_r\}_{r=1}^{n_u}$ is their geometric mean $d^L=\exp(n_u^{-1}\sum_r\log d_r)$, regressed as $\log(1+d^L)$.
Let $\mathcal F_v$ collect the (latent frame, valid cell, target) triples $(\widetilde{\mathbf x},u,d^L)$ of view $v$ in a fitting minibatch.
The fitting loss is
\begin{equation}
    \mathcal L_{\mathrm{head}}(\phi)
      =\sum_{v\in\mathcal V}\frac{1}{|\mathcal F_v|}
       \sum_{(\widetilde{\mathbf x},u,d^L)\in\mathcal F_v}
       \rho_\beta\!\left(
          \left|[\Pi_\phi(\widetilde{\mathbf x})](u)-\log(1+d^L)\right|
       \right),
    \label{eq:app:head_fit}
\end{equation}
where $\rho_\beta$ is the Huber penalty of Sec.~\ref{sec:method:cpp} with transition $\beta$ in log-depth units.
After fitting, $\phi$ is fixed: CPP gradients pass through $\Pi_\phi$ into the generated latents, and the head cannot absorb geometric error by adapting to the generator.

\noindent\textbf{Correspondence and anchors.}
CPP compares two 3D points per LiDAR-supported cell, the generated $\widehat{\mathbf P}_{v,t}(u)$ and the measured $\mathbf P^\star_{v,t}(u)$ of Eq.~(\ref{eq:method:cpp_geometry}).
Both lie on the ray $\mathbf b_v(u)$ and are placed in $\mathcal C_0$ by $\mathbf E_v$ followed by the generated chain $\widehat{\mathbf T}_t$ or the recorded pose $\mathbf T^\star_t$, so correspondence is given by the index $(v,t,u)$ and no point-cloud registration is needed.
The anchors $\mathcal A\subseteq\{\kappa(j):j>0\}$ are future frames: at $t=0$ the transform is the identity and the comparison would supervise depth alone.
A cell is valid when it holds at least $n_{\min}$ LiDAR returns, its pooled depth is finite and lies in $(0,d_{\max}]$, and its predicted depth is finite; invalid cells receive zero weight and zero gradient.

\noindent\textbf{Reduction.}
For example $b$ and view $v$, let $n_{b,v}$ count the valid cells over all anchors and $\mathcal L^{(b,v)}_{\mathrm{cpp}}$ be their mean Huber error.
With $\mathcal B_+=\{b:\sum_v n_{b,v}>0\}$, the noise-weighted minibatch loss is
\begin{equation}
    \overline{\mathcal L}_{\mathrm{cpp}}
       =\frac{1}{|\mathcal B_+|}
        \sum_{b\in\mathcal B_+}w(\sigma_b)
        \frac{\sum_v n_{b,v}\mathcal L_{\mathrm{cpp}}^{(b,v)}}
             {\sum_v n_{b,v}}.
    \label{eq:app:cpp_reduction}
\end{equation}
Each example averages over all of its valid correspondences, so a densely observed view carries proportionally more weight than a sparsely observed one, and examples without LiDAR support are left out of the mean rather than diluting it.

\noindent\textbf{Gradients to depth and motion.}
Let $\mathbf M=\widehat{\mathbf T}_t\mathbf E_v$ with rotation $\mathbf R_{\mathbf M}$, let $\mathbf r=\widehat{\mathbf P}_{v,t}(u)-\mathbf P^\star_{v,t}(u)$ be the residual with $n=\|\mathbf r\|_2$, and let $\mathbf g=\rho'_\delta(n)\,\mathbf r/n$ with $\rho'_\delta(n)=\min(n/\delta,1)$.
For one correspondence, the Huber term $\rho_\delta(n)$ has the gradients
\begin{equation}
    \frac{\partial\rho_\delta}{\partial\widehat d}
      =\mathbf g^{\!\top}\mathbf R_{\mathbf M}\frac{\mathbf b_v(u)}{b_{v,z}(u)},
    \qquad
    \frac{\partial\rho_\delta}{\partial\widehat{\mathbf t}_t}
      =\mathbf g,
    \qquad
    \frac{\partial\rho_\delta}{\partial\boldsymbol\theta_t}
      =\big(\widehat{\mathbf P}_{v,t}(u)-\widehat{\mathbf t}_t\big)\times\mathbf g,
    \label{eq:app:cpp_gradients}
\end{equation}
where $\widehat d=\widehat d^v_{j(t)}(u)$, $\widehat{\mathbf t}_t$ is the translation of $\widehat{\mathbf T}_t$, and $\boldsymbol\theta_t$ is a small rotation of the generated camera about its own centre $\widehat{\mathbf t}_t$, expressed in $\mathcal C_0$.
The generated depth is thus corrected along its transformed viewing ray, by the component of the residual in that direction, while the generated pose receives the residuals of all cells at the anchor as a net force on its translation and a net torque about the camera centre on its rotation.
The pose gradient reaches every relative motion up to $t$ through the composition $\widehat{\mathbf T}_t=\boldsymbol\Delta(\widehat{\mathbf a}_1)\cdots\boldsymbol\Delta(\widehat{\mathbf a}_t)$, and the depth gradient reaches the latent $\widehat{\mathbf x}_{v,D}$ through $\Pi_\phi$; both then reach the velocity outputs through Eq.~(\ref{eq:method:clean_estimate}), with the factor $-\sigma$.
One residual vector drives both branches, which is the coupling of Sec.~\ref{sec:method:cpp} in explicit form.

\noindent\textbf{Relation to decoupled supervision.}
Write $\mathbf p(d)=\mathbf E_v\,d\,\mathbf b_v(u)/b_{v,z}(u)$ for the point at depth $d$ on the ray of cell $u$, so that $\widehat{\mathbf P}_{v,t}(u)=\widehat{\mathbf T}_t\,\mathbf p(\widehat d)$ and $\mathbf P^\star_{v,t}(u)=\mathbf T^\star_t\,\mathbf p(d^L)$.
The residual of one correspondence, $\mathbf r=\widehat{\mathbf P}_{v,t}(u)-\mathbf P^\star_{v,t}(u)$, is the vector whose Huber norm CPP penalizes.
It splits exactly into three parts,
\begin{equation}
    \mathbf r
      =\underbrace{\mathbf T^\star_t\,\mathbf p(\widehat d)-\mathbf T^\star_t\,\mathbf p(d^L)}_{\mathbf r_D}
      +\underbrace{\widehat{\mathbf T}_t\,\mathbf p(d^L)-\mathbf T^\star_t\,\mathbf p(d^L)}_{\mathbf r_T}
      +\underbrace{\big(\widehat{\mathbf T}_t-\mathbf T^\star_t\big)\big(\mathbf p(\widehat d)-\mathbf p(d^L)\big)}_{\mathbf r_{DT}}.
    \label{eq:app:cpp_split}
\end{equation}
The first part, $\mathbf r_D=(\widehat d-d^L)\,\mathbf R^\star_t\mathbf R_{\mathbf E_v}\mathbf b_v(u)/b_{v,z}(u)$, is the depth error carried along the recorded transformed ray; it depends on the generated depth only.
The second, $\mathbf r_T$, applies the generated pose to the measured point and compares with the recorded pose; it depends on the generated motion only.
The third, $\mathbf r_{DT}$, is the product of the two errors.

A decoupled objective would penalize the two errors separately, as $\rho_\delta(\|\mathbf r_D\|)+\rho_\delta(\|\mathbf r_T\|)$.
The first is a Huber loss on the metric depth error at the LiDAR-supported cells, up to the ray obliquity $\|\mathbf b_v(u)\|/b_{v,z}(u)$; it is related to, but not the same as, the depth stream's flow-matching loss, which acts in latent space on a dense label anchored to the same LiDAR sweep (App.~\ref{app:data:depth}).
The second is a loss on the composed pose $\widehat{\mathbf T}_t$ evaluated on the measured points, with the translation error entering directly and the rotation error scaled by the distance of the point; the hinges (App.~\ref{app:method:hinges}) act on the same composed positions but supervise feasibility margins rather than agreement with the recorded pose.

CPP applies $\rho_\delta$ to the sum instead, and this changes the gradients.
With $\mathbf g$ the Huber-scaled residual of Eq.~(\ref{eq:app:cpp_gradients}), the depth gradient $\mathbf g^{\!\top}\mathbf R_{\mathbf M}\mathbf b_v(u)/b_{v,z}(u)$ contains the components of $\mathbf r_T$ and $\mathbf r_{DT}$ along the ray, so the depth is moved by the motion error.
The pose gradients $\mathbf g$ and $(\widehat{\mathbf P}_{v,t}(u)-\widehat{\mathbf t}_t)\times\mathbf g$ contain $\mathbf r_D$, so the motion is moved by the depth error.
The two branches share one residual and one Huber scale: each is corrected toward agreement with the other's current prediction as well as with the measurement, and a cell whose joint error is large drives both.
The ablation of Sec.~\ref{sec:exp:ablations} measures the benefit of the added objective, 1.0 PDMS over flow matching with the hinges and 2.1 over flow matching alone; a matched comparison against the decoupled objective above would separate the contribution of the coupling itself.

\subsection{Geometric Hinge Losses}
\label{app:method:hinges}

CPP supervises the generated depth--motion pair; the two hinges supervise waypoint positions through the generated motion alone, independently of generated depth.
Both hinges act on the planar waypoints $\widehat{\mathbf p}_t=\pi_{xz}(\operatorname{trans}(\widehat{\mathbf T}_t))$, $t=1,\ldots,T$, in the $(x,z)$ plane of $\mathcal C_0$, which serves as the ground plane: $x$ points right and $z$ forward from the front camera at the current time, and height is discarded.
Labels observed at a future time $t$ are placed in this plane with the recorded pose $\mathbf T^\star_t$, as the measured points of CPP are, so waypoints and labels are compared in one frame.
Both margins apply to positions rather than heading-dependent vehicle footprints, and the obstacle and map targets supply training supervision only.

\noindent\textbf{Obstacle labels.}
For each future time $t$, the static set $\mathcal S_t\subset\mathbb R^2$ holds the planar positions of LiDAR returns from the sweep at $t$ that belong to unannotated structure at vehicle height, such as curbs, poles, and barriers, and the box set $\mathcal O_t$ holds the footprints of the objects annotated at $t$, each an oriented rectangle $B=(\mathbf c_B,\mathbf h_B,\psi_B)$ with centre $\mathbf c_B\in\mathbb R^2$, half-extents $\mathbf h_B=(\tfrac12L_B,\tfrac12W_B)$ along its length and width, and yaw $\psi_B$.
Neither set contains the ego vehicle; App.~\ref{app:data} describes their construction.

\noindent\textbf{Signed distance to an oriented rectangle.}
For a position $\mathbf p$, let $\mathbf q=\mathbf R_2(-\psi_B)(\mathbf p-\mathbf c_B)$ be its coordinates along the length and width axes of $B$, where $\mathbf R_2(\psi)$ is the planar rotation by $\psi$ and $\psi_B$ is the angle from the $x$ axis to the length axis, and let $\boldsymbol\zeta_B(\mathbf p)=|\mathbf q|-\mathbf h_B=(\zeta_{B,1},\zeta_{B,2})$ be the elementwise excess of $|\mathbf q|$ over the half-extents.
The signed distance to the boundary of $B$ is
\begin{equation}
    \operatorname{sd}(\mathbf p,B)
       =\big\|\operatorname{ReLU}(\boldsymbol\zeta_B(\mathbf p))\big\|_2
        +\min\!\big(\max(\zeta_{B,1}(\mathbf p),\zeta_{B,2}(\mathbf p)),\,0\big).
    \label{eq:app:rectangle_sdf}
\end{equation}
Outside the rectangle at least one component of $\boldsymbol\zeta_B$ is positive, the second term vanishes, and the first is the Euclidean distance to the nearest edge or corner.
Inside, both components are negative, the first term vanishes, and the second is minus the distance to the nearest edge.
The function is continuous across the boundary, and its gradient has unit norm almost everywhere and points away from the rectangle on both sides of the boundary.
An unsigned distance would instead grow towards the centre of the footprint, and the hinge would drive a penetrating waypoint deeper into it.

\noindent\textbf{Obstacle clearance.}
For a position $\mathbf p$, the clearance at time $t$ is
\begin{equation}
    d_t(\mathbf p)=\min\!\left\{
       \min_{\mathbf s\in\mathcal S_t}\|\mathbf p-\mathbf s\|_2,\ 
       \min_{B\in\mathcal O_t}\operatorname{sd}(\mathbf p,B)
    \right\},
    \label{eq:app:clearance}
\end{equation}
where an empty set does not contribute and $d_t\equiv+\infty$ when $\mathcal S_t$ and $\mathcal O_t$ are both empty, so that the term vanishes; time-indexed annotations follow other vehicles' recorded locations.
The recorded clearance $d_t^\star=d_t(\mathbf p^\star_t)$ evaluates it at the recorded waypoint $\mathbf p^\star_t=\pi_{xz}(\operatorname{trans}(\mathbf T^\star_t))$ and is negative when $\mathbf p^\star_t$ lies inside a footprint.
Capping the margin at $m_t=\min(m_{\mathrm{obs}},d_t^\star)$ makes the recorded trajectory incur zero loss at every $t$: in free space the hinge requires the nominal margin, and where the recorded path passed closer, only the clearance that path kept.
For the nominal margin $m_{\mathrm{obs}}>0$, the loss is
\begin{equation}
    \mathcal L_{\mathrm{obs}}
      =\frac{1}{T}\sum_{t=1}^T
        \big[\operatorname{ReLU}\!\big(m_t-d_t(\widehat{\mathbf p}_t)\big)\big]^2,
    \label{eq:app:obstacle_loss}
\end{equation}
with $\operatorname{ReLU}(x)=\max(x,0)$, so only clearance below $m_t$ is penalized.
The gradient of the penalty $[\operatorname{ReLU}(m_t-d_t(\widehat{\mathbf p}_t))]^2$ with respect to $\widehat{\mathbf p}_t$ is $-2\operatorname{ReLU}(m_t-d_t(\widehat{\mathbf p}_t))\,\nabla d_t(\widehat{\mathbf p}_t)$, where $\nabla d_t(\widehat{\mathbf p}_t)$ is the unit vector to $\widehat{\mathbf p}_t$ from the nearest obstacle point (a static return, or the closest boundary point of a footprint, which near a corner is the corner itself) and, inside a footprint, the outward normal of its nearest edge.
A violating waypoint is thus moved directly away from the nearest obstacle with a force proportional to the deficit, and the gradient vanishes once the margin is met.

\noindent\textbf{Drivable-area label.}
The drivable region $\mathcal R$ is the union of the map polygons on which driving is permitted (road blocks, intersections, and car-park areas).
Its signed distance field is stored on a raster $\mathbf S\in\mathbb R^{N_z\times N_x}$ of square cells with spacing $r_S$, covering $[x_{\min},x_{\min}+r_SN_x)\times[z_{\min},z_{\min}+r_SN_z)$ in the ground plane of $\mathcal C_0$, a region around the current camera position that extends farther forward than backward.
The entry $S_{i_z,i_x}$ is the Euclidean distance in metres from the cell centre $\big(x_{\min}+(i_x+\tfrac12)r_S,\ z_{\min}+(i_z+\tfrac12)r_S\big)$ to the boundary of $\mathcal R$, positive inside $\mathcal R$ and negative outside.

\noindent\textbf{Drivable-area clearance.}
With $S(\mathbf p)$ the signed distance to the drivable boundary, positive inside the drivable region, and the required interior clearance $m_{\mathrm{drv}}>0$,
\begin{equation}
    \mathcal L_{\mathrm{drv}}
       =\frac{1}{T}\sum_{t=1}^T
         \big[\operatorname{ReLU}\!\big(m_{\mathrm{drv}}-S(\widehat{\mathbf p}_t)\big)\big]^2,
    \label{eq:app:drivable_loss}
\end{equation}
which penalizes waypoints near or outside the drivable boundary and vanishes beyond the required interior margin.
The value $S(\mathbf p)$ is the bilinear interpolation of the raster at $\mathbf p$.
The continuous cell indices
\begin{equation}
    \iota_x(\mathbf p)=\frac{p_x-x_{\min}}{r_S}-\tfrac12,
    \qquad
    \iota_z(\mathbf p)=\frac{p_z-z_{\min}}{r_S}-\tfrac12
    \label{eq:app:sdf_index}
\end{equation}
take integer values exactly at cell centres.
With $i_x=\lfloor\iota_x\rfloor$, $a_x=\iota_x-i_x$, and likewise $i_z$, $a_z$,
\begin{equation}
    S(\mathbf p)
      =(1-a_z)(1-a_x)\,S_{i_z,i_x}
      +(1-a_z)\,a_x\,S_{i_z,i_x+1}
      +a_z(1-a_x)\,S_{i_z+1,i_x}
      +a_za_x\,S_{i_z+1,i_x+1},
    \label{eq:app:sdf_bilinear}
\end{equation}
where indices outside the raster are clamped to its border, so a waypoint beyond the stored region takes the value of the nearest border cell.
$S(\mathbf p)$ is continuous and piecewise bilinear, and its gradient, obtained from the interpolation weights, approximates the unit inward normal of the drivable boundary.
The drivable penalty has the gradient of the obstacle case with $S$ and $m_{\mathrm{drv}}$ in place of $d_t$ and $m_t$, so a violating waypoint moves along $\nabla S$, towards the interior; beyond the raster, the clamped coordinate receives no gradient.

\noindent\textbf{Reduction.}
Each hinge averages its squared penalty over the $T$ future waypoints of an example, is weighted by $w(\sigma)$, and is averaged over all examples of the minibatch; unlike CPP, every example contributes, since the labels are always available.
Although only positions are penalized, $\widehat{\mathbf p}_t$ is the endpoint of the composed chain and depends on the translations of $\widehat{\mathbf a}_1,\ldots,\widehat{\mathbf a}_t$ and the rotations of $\widehat{\mathbf a}_1,\ldots,\widehat{\mathbf a}_{t-1}$; a heading error in $\widehat{\mathbf a}_{t'}$ is therefore corrected through every later waypoint it displaces.
As for CPP, the gradients reach the action velocity output through Eq.~(\ref{eq:method:clean_estimate}) with the factor $-\sigma$.

\subsection{Output-Gradient Balancing}
\label{app:method:balancing}

The geometric losses differ from the flow-matching losses in units and sparsity, and their raw gradients at the velocity outputs can differ from the flow gradients by orders of magnitude.
Fixed weights would have to follow these magnitudes as training changes them; the balancing of Sec.~\ref{sec:method:training_inference} instead recomputes the coefficients on every update from the gradients themselves.
For target fractions $\eta_k$, each active geometric term contributes an $\eta_k$ fraction of the action flow-gradient norm at the action output, and a gain $\gamma$ on CPP's depth branch imposes the analogous ratio at the depth output; all gradients are measured before scaling, $\operatorname{sg}$ holds the coefficients fixed during backpropagation, and zero-denominator ratios are set to zero.

\noindent\textbf{Gradients at the outputs.}
On the current minibatch, let the unweighted gradients, with the depth-side gain $\gamma$ held at one, be
\begin{equation}
\begin{aligned}
    \mathbf g_A&=\nabla_{\widehat{\mathbf v}_A}\mathcal L_{\mathrm{FM},A},
    &\mathbf g_k^A&=\nabla_{\widehat{\mathbf v}_A}\overline{\mathcal L}_k,\\
    \mathbf g_D&=\nabla_{\widehat{\mathbf v}_D}\mathcal L_{\mathrm{FM},V},
    &\mathbf g_{\mathrm{cpp}}^D&=\nabla_{\widehat{\mathbf v}_D}\overline{\mathcal L}_{\mathrm{cpp}},
\end{aligned}
    \label{eq:app:output_gradients}
\end{equation}
where $k\in\mathcal K$ and $\widehat{\mathbf v}_D$ collects the depth velocity outputs of all views, so that the visual flow gradient is restricted to depth for the depth-side comparison.
Each vector concatenates the representation-space outputs of the whole minibatch; transformer hidden states are not involved.
With the norms $\nu_A=\|\mathbf g_A\|_2$, $\nu_k^A=\|\mathbf g_k^A\|_2$, $\nu_D=\|\mathbf g_D\|_2$, and $\nu_{\mathrm{cpp}}^D=\|\mathbf g_{\mathrm{cpp}}^D\|_2$, the action coefficients read
\begin{equation}
    s_k=\operatorname{sg}\!\left(\eta_k\frac{\nu_A}{\nu_k^A}\right),
    \label{eq:app:scales}
\end{equation}
and in the final backward pass the action output receives
\begin{equation}
    \nabla_{\widehat{\mathbf v}_A}\mathcal L_{\mathrm{total}}
       =\lambda_A\Big(\mathbf g_A+\sum_{k\in\mathcal K}s_k\mathbf g_k^A\Big),
    \qquad
    \|\lambda_As_k\mathbf g_k^A\|_2
       =\eta_k\|\lambda_A\mathbf g_A\|_2
    \label{eq:app:ratio_identities}
\end{equation}
for every active term.
Each geometric term pushes the action output with $\eta_k$ times the norm of the action flow push; because $s_k$ is a constant in the backward pass, the identity holds exactly, at minibatch level, on the update for which $s_k$ is computed.

\noindent\textbf{Depth-side gain.}
CPP also reaches the depth output, and $s_{\mathrm{cpp}}$ alone does not balance it there.
Substituting $s_{\mathrm{cpp}}$, the ratio it would induce at the depth output is
\begin{equation}
    \frac{\|\lambda_As_{\mathrm{cpp}}\mathbf g^D_{\mathrm{cpp}}\|_2}{\|\lambda_V\mathbf g_D\|_2}
    =\eta_{\mathrm{cpp}}\,
     \frac{\lambda_A\nu_A}{\lambda_V\nu_D}\,
     \frac{\nu^D_{\mathrm{cpp}}}{\nu^A_{\mathrm{cpp}}},
    \label{eq:app:induced_ratio}
\end{equation}
which equals $\eta_{\mathrm{cpp}}$ only when CPP divides its gradient between the depth and action outputs in the same proportion as the weighted flow losses divide theirs; nothing in the objective ties the two proportions.
A second coefficient is therefore applied on the depth branch of CPP alone, through the operator
\begin{equation}
    \mathcal G_\gamma(\widehat d)
       =\operatorname{sg}(\widehat d)
        +\gamma\big(\widehat d-\operatorname{sg}(\widehat d)\big),
    \label{eq:app:depth_gain}
\end{equation}
which returns $\widehat d$ unchanged in the forward pass and multiplies the gradient flowing back through it by $\gamma$; the value of CPP, its motion branch, and the flow-matching losses are unaffected.
With the gain, the depth output receives
\begin{equation}
    \nabla_{\widehat{\mathbf v}_D}\mathcal L_{\mathrm{total}}
       =\lambda_V\mathbf g_D+\lambda_As_{\mathrm{cpp}}\gamma\,\mathbf g_{\mathrm{cpp}}^D,
    \qquad
    \|\lambda_As_{\mathrm{cpp}}\gamma\,\mathbf g_{\mathrm{cpp}}^D\|_2
       =\eta_{\mathrm{cpp}}\|\lambda_V\mathbf g_D\|_2,
    \label{eq:app:depth_identity}
\end{equation}
and solving the identity for the gain gives
\begin{equation}
    \gamma=\operatorname{sg}\!\left(
      \frac{\eta_{\mathrm{cpp}}\lambda_V
        \|\nabla_{\widehat{\mathbf v}_D}\mathcal L_{\mathrm{FM},V}\|_2}
      {\lambda_A s_{\mathrm{cpp}}
        \|\nabla_{\widehat{\mathbf v}_D}\overline{\mathcal L}_{\mathrm{cpp}}\|_2}
      \right),
    \label{eq:app:gamma}
\end{equation}
whose factor $\lambda_V/\lambda_A$ appears because CPP is weighted by $\lambda_A$ in $\mathcal L_{\mathrm{total}}$ while its depth-side reference, the depth part of the visual flow loss, is weighted by $\lambda_V$.
In terms of the norms, $\gamma=\lambda_V\nu_D\nu^A_{\mathrm{cpp}}/(\lambda_A\nu_A\nu^D_{\mathrm{cpp}})$, the reciprocal of the last two factors in Eq.~(\ref{eq:app:induced_ratio}), and does not depend on $\eta_{\mathrm{cpp}}$.

\noindent\textbf{The depth backward path.}
Between the CPP value and the depth velocity output of a view, the gradient passes through the unprojection, the exponential and clip of the depth head (Sec.~\ref{sec:method:cpp}), the fixed head, and the clean estimate of Eq.~(\ref{eq:method:clean_estimate}).
Writing $\ell=\Pi_\phi(\widehat{\mathbf x}_{v,D})$ for the head's log-depth output and $\mathbf J_\Pi$ for its Jacobian, the view-$v$ block of $\mathbf g^D_{\mathrm{cpp}}$ is
\begin{equation}
    \mathbf g^{D}_{\mathrm{cpp},v}
      =-\sigma\,\mathbf J_\Pi^{\!\top}
        \operatorname{diag}\!\big(\mathbf 1[\ell_{\min}<\ell<\ell_{\max}]\,\big(1+\widehat d\big)\big)\,
        \nabla_{\widehat d}\overline{\mathcal L}_{\mathrm{cpp}},
    \label{eq:app:depth_backward}
\end{equation}
where $\nabla_{\widehat d}\overline{\mathcal L}_{\mathrm{cpp}}$ collects the per-cell terms of Eq.~(\ref{eq:app:cpp_gradients}) with the weights of Eq.~(\ref{eq:app:cpp_reduction}) and is zero at invalid cells and at latent frames $j$ with $\kappa(j)\notin\mathcal A$, $1+\widehat d$ is the derivative of the exponential, the indicator $\mathbf 1[\cdot]$, equal to one inside the clip range and zero outside, is the derivative of the clip, and $-\sigma$ is the derivative of the clean estimate with respect to the velocity; cells whose log-depth leaves the clip range send no gradient.
The final backward pass multiplies this vector by $\lambda_As_{\mathrm{cpp}}\gamma$, the gain entering at $\widehat d$ through $\mathcal G_\gamma$.
Because backpropagation is linear in the gradient it carries, a scalar gain anywhere on the segment from the unprojected points $\mathcal U_v(\widehat d,u)$ back to $\widehat{\mathbf x}_{v,D}$, which only the depth branch traverses, has this effect; a scale on the loss would also reach the motion branch, and one at $\widehat{\mathbf v}_{v,D}$ would also scale the flow gradient added there.
Since $\nu^D_{\mathrm{cpp}}$ is measured with $\gamma=1$, it already includes the effect of the head, the exponential, and the clip, so Eq.~(\ref{eq:app:depth_identity}) holds exactly without any separate correction for these factors.

\noindent\textbf{One training update.}
Algorithm~\ref{alg:app:training} lists one update.
The norms are obtained by differentiating each loss only as far as the velocity outputs, at a cost independent of the transformer, and are formed over the whole minibatch, so the coefficients are common to all of it.
A coefficient whose denominator vanishes is set to zero, so an inactive term drops out of the update.
The gain is measured at $\gamma=1$ and set before the single backward pass through the model, which is the only pass that reaches the parameters.

\begin{algorithm}[t]
    \caption{One \method{} training update with output-gradient balancing}
    \label{alg:app:training}
    \small
    \begin{algorithmic}[1]
        \Require minibatch of examples with images $\mathbf I^v_{0:T}$, depth $\mathbf D^v_{0:T}$, motion $\mathbf a_{0:T}$, context $c$, pooled LiDAR depths, recorded poses $\mathbf T^\star_t$, obstacle sets $\mathcal S_t,\mathcal O_t$, and raster $\mathbf S$; parameters $\theta$; fixed $\mathcal E$ and $\Pi_\phi$; weights $\lambda_V,\lambda_A$; target fractions $\eta_k$
        \State $\mathbf x_{v,I}\gets\mathcal E(\mathbf I^v_{0:T})$,\ $\mathbf x_{v,D}\gets\mathcal E(\mathcal C(\mathbf D^v_{0:T}))$,\ $\mathbf x_A\gets\mathbf a_{0:T}$ \Comment{clean representations}
        \State draw $\sigma_b$ per example and $\boldsymbol\epsilon$; form $\mathbf x^\sigma$ and $\mathbf v^\star$ on generated entries \Comment{Eq.~(\ref{eq:method:flow_path})}
        \State $\widehat{\mathbf v}\gets v_\theta(\mathbf x^\sigma,\sigma;c)$ \Comment{one forward pass}
        \State $\widehat{\mathbf x}_{v,D}\gets\mathbf x^\sigma_{v,D}-\sigma_b\widehat{\mathbf v}_{v,D}$,\ $\widehat{\mathbf a}_t\gets\mathbf a^\sigma_t-\sigma_b\widehat{\mathbf v}_{A,t}$ \Comment{Eq.~(\ref{eq:method:clean_estimate})}
        \State $\widehat d\gets\exp(\operatorname{clip}(\Pi_\phi(\widehat{\mathbf x}_{v,D}),\ell_{\min},\ell_{\max}))-1$;\ $\widehat{\mathbf T}_t\gets\boldsymbol\Delta(\widehat{\mathbf a}_1)\cdots\boldsymbol\Delta(\widehat{\mathbf a}_t)$;\ $\widehat{\mathbf p}_t\gets\pi_{xz}(\operatorname{trans}(\widehat{\mathbf T}_t))$
        \State $\mathcal L_{\mathrm{FM},V},\mathcal L_{\mathrm{FM},A}$ by Eq.~(\ref{eq:method:native_losses});\ $\overline{\mathcal L}_{\mathrm{cpp}}$ by Eq.~(\ref{eq:app:cpp_reduction}) with $\mathcal G_{\gamma=1}(\widehat d)$;\ $\overline{\mathcal L}_{\mathrm{obs}},\overline{\mathcal L}_{\mathrm{drv}}$ by Sec.~\ref{app:method:hinges}
        \State $\mathbf g_A,\mathbf g_k^A,\mathbf g_D,\mathbf g_{\mathrm{cpp}}^D$ by differentiation to the velocity outputs \Comment{Eq.~(\ref{eq:app:output_gradients})}
        \State $\nu_A,\nu_k^A,\nu_D,\nu_{\mathrm{cpp}}^D\gets$ minibatch norms of these gradients
        \State $s_k$ by Eq.~(\ref{eq:app:scales}), $\gamma$ by Eq.~(\ref{eq:app:gamma}); set $\gamma$ in $\mathcal G_\gamma$ \Comment{constants in the backward pass}
        \State $\mathcal L_{\mathrm{total}}\gets\lambda_V\mathcal L_{\mathrm{FM},V}+\lambda_A\big(\mathcal L_{\mathrm{FM},A}+\sum_{k\in\mathcal K}s_k\overline{\mathcal L}_k\big)$ \Comment{Eq.~(\ref{eq:method:total_objective})}
        \State backpropagate $\mathcal L_{\mathrm{total}}$ through $v_\theta$ once and update $\theta$
    \end{algorithmic}
\end{algorithm}

\subsection{Inference and Trajectory Selection}
\label{app:method:inference}

\noindent\textbf{Joint denoising and decoding.}
At inference, all generated representations are initialized independently from Gaussian noise and denoised jointly with the UniPC sampler~\cite{unipc} of the pretrained framework~\cite{cosmos3}.
Current RGB, motion history, and text stay fixed at every model evaluation, and the padding channels of the action stream stay zero.
The VAE decoder $\mathcal D$ reconstructs RGB and depth at full resolution; averaging the decoded depth channels and applying Eq.~(\ref{eq:app:depth_inverse}) gives metric depth.
The depth head $\Pi_\phi$ provides geometric supervision during training only and is not used at inference.

\noindent\textbf{Ego trajectory and selection.}
The generated pose rows are converted to relative rotations through the continuous representation~\cite{rotation6d} and composed into $\widehat{\mathbf T}_t$.
Conjugation by the fixed front-camera calibration gives $\widehat{\mathbf T}^{\mathrm{ego}}_t=\mathbf G_{\mathrm F}\widehat{\mathbf T}_t\mathbf G_{\mathrm F}^{-1}$.
With $\widehat{\mathbf R}^{\mathrm{ego}}_t$ its rotation, the first two translation coordinates give the planar ego position and the yaw is $\operatorname{atan2}([\widehat{\mathbf R}^{\mathrm{ego}}_t]_{21},[\widehat{\mathbf R}^{\mathrm{ego}}_t]_{11})$.
For $K$ joint samples, let $\mathbf y_t^{(k)}$ be the planar ego position at time $t$ in candidate $k$.
We select
\begin{equation}
    k^*=\operatorname*{arg\,min}_{k\in\{1,\ldots,K\}}
          \sum_{l=1}^{K}\frac{1}{T}\sum_{t=1}^{T}
          \|\mathbf y_t^{(k)}-\mathbf y_t^{(l)}\|_2.
    \label{eq:app:medoid}
\end{equation}
The score uses planar positions only; yaw, generated geometry, measured targets, and simulator feedback play no part.
The selected candidate is returned in full, including its yaw, without averaging across candidates or any further tie-break, and its video and depth are the outputs of the same joint sample.

%% file: sections/appendix_data.tex

\section{Data and Labels}
\label{app:data}

\method{} is trained on recorded driving logs.
Every supervision signal is derived from the sensors, calibration, poses, object annotations, and maps that the logs already contain without any additional manual annotation.
This appendix describes the training windows (Sec.~\ref{app:data:windows}), the dense metric depth labels of the depth streams (Sec.~\ref{app:data:depth}), the LiDAR targets of CPP (Sec.~\ref{app:data:lidar}), and the obstacle and drivable-area labels of the hinge losses (Sec.~\ref{app:data:hinges}).
All geometric labels of a window are expressed in $\mathcal C_0$, the front-camera frame at the current time, using the recorded poses, so that generated depth, generated motion, and every label share one frame.

\subsection{Source Data and Training Windows}
\label{app:data:windows}

\noindent\textbf{Logs and split.}
We use the \texttt{navtrain} split of NAVSIM~\cite{navsim}, built on OpenScene~\cite{openscene} and nuPlan~\cite{nuplan}: 1{,}192 logs recorded at 2\,Hz.
Each log frame provides eight camera images, one merged LiDAR sweep, the ego pose, the driving command, and oriented 3D boxes with track identities for all annotated objects.
We use the three forward-facing cameras: front, front-left, and front-right.
Camera intrinsics, Brown lens distortion, and camera-to-LiDAR extrinsics are constant per vehicle and differ between vehicles; images are used as recorded, without undistortion, and every projection in the pipeline uses the vehicle's own model.

\noindent\textbf{Windows.}
One training window is formed at every \texttt{navtrain} scene token: the current frame, the eight following frames ($T=8$, 4\,s), and the preceding frame, which supplies the motion-history row $\mathbf a_0$.
A total of 103{,}281 windows are used for training.
Consecutive tokens of a log are mostly one frame apart, so windows overlap heavily; no command-based filtering or resampling is applied.
The front view is used at $832\times468$ and the side views at $416\times234$.

\noindent\textbf{Motion and text.}
The pose of the front camera at each frame is the composition of the recorded ego pose with the camera extrinsics.
The relative motion $\boldsymbol\Delta_t$ of Sec.~\ref{sec:method:preliminaries} is the pose of frame $t$ expressed in the front-camera frame of frame $t-1$, stored as translation in metres and the first two columns of the rotation matrix.
The text context is a fixed template that names the city, the discrete driving command (turn left, go straight, turn right, or follow the route), and the current speed; it is replaced by the empty string in 10\% of training examples to support classifier-free guidance.

\subsection{Dense Metric Depth Labels}
\label{app:data:depth}

The depth streams are generated modalities and are supervised through the flow-matching loss like RGB, which requires a dense depth video for every view and frame.
LiDAR alone is too sparse for this purpose, so the labels are metric depth predictions of a feed-forward reconstruction model anchored to the LiDAR sweep of each frame.
The labels serve the flow-matching loss only; CPP compares generated depth with the LiDAR itself (Sec.~\ref{app:data:lidar}).

\noindent\textbf{Reconstruction with a LiDAR prior.}
For each frame of a window, MapAnything~\cite{mapanything} is run once on the three views jointly, with metric scale enabled and given the calibrated camera poses and the frame's LiDAR sweep projected into each view as a sparse metric depth prior.
The predicted metric depth and validity mask are resampled to the image grid of Sec.~\ref{app:data:windows}, so that the depth label, the LiDAR targets, and the calibrated rays share one pixel grid.
Sky pixels are identified by the sky probability of Depth Anything~3~\cite{da3} and set to $d_{\max}$.

\noindent\textbf{Hole filling.}
MapAnything leaves about one fifth of the non-sky pixels of a frame invalid, concentrated in the band around the horizon.
Structure for these holes is taken from the monocular metric prediction of Depth Anything~3 \cite{da3} on the same frames, scaled to the LiDAR once per window.
Rather than directly use this prediction, we correct it towards the label.
Where both are valid, we compute the log-ratio between the label and the monocular depth; inside a hole, this ratio is interpolated smoothly from the surrounding valid pixels, as the solution of Laplace's equation \cite{poisson_editing} with those pixels as boundary values, and the fill is the monocular depth scaled by the interpolated ratio.
Holes that the monocular model also leaves unstructured are filled by interpolating the label's own log depth in the same way, and are assigned $d_{\max}$ only where the interpolated depth exceeds 40\,m.
On the front view, 65\% of the pixels of the final label come from MapAnything, 16\% from the corrected fill, 15\% are sky, and 4\% are smooth extensions; the two side views are similar.

\noindent\textbf{Bias correction.}
Compared with the LiDAR, MapAnything reads about 3\% far at 10--25\,m and 5\% near at 60--80\,m, a bias that is a function of depth alone and is consistent across vehicles.
We fit a piecewise-linear correction of log depth, with the range from 2\,m to 80\,m partitioned into ten segments, on 34.6\,M label--LiDAR pixel pairs from all sixteen vehicles, keeping 30\% of the driving logs aside to check the fit, and apply it to every label before encoding.
After correction, the median log-ratio between label and LiDAR is within $0.023$ in every depth band.

\noindent\textbf{Encoding.}
The corrected label is clipped to $[d_{\min},d_{\max}]=[0.5,80]$\,m, encoded by Eq.~(\ref{eq:app:depth_codec}), and stored as an 8-bit video at the resolution of the RGB stream; the 255 levels span the log-depth range in steps of about 1.6\% relative depth.

\noindent\textbf{Verification.}
Table~\ref{tab:app:label_quality} compares the final labels with the raw LiDAR on 1{,}000 windows covering all sixteen vehicles, using a projection of the sweeps written independently of the label pipeline.
Agreement is measured on the LiDAR-supported latent cells that CPP uses and on all pixels with a LiDAR return within 25\,m; the front--side seam is checked on the columns where the views overlap.

\begin{table}[!ht]
    \centering
    \caption{\textbf{Depth-label quality against LiDAR} on 1{,}000 training windows. AbsRel is the mean absolute relative depth error; the cell-level column uses the geometric-mean LiDAR depth of latent cells with at least three returns, the pixel-level column all LiDAR returns within 25\,m. Sky error is the fraction of pixels labelled sky that carry a LiDAR return.}
    \label{tab:app:label_quality}
    \small
    \setlength{\tabcolsep}{6pt}
    \begin{tabular}{@{}lccc@{}}
        \toprule
        \textbf{View} & \textbf{AbsRel, cells} & \textbf{AbsRel, pixels $\le 25$\,m} & \textbf{Sky error} \\
        \midrule
        Front       & 0.041 & 0.073 & 0.05\% \\
        Front-left  & 0.051 & 0.113 & 0.22\% \\
        Front-right & 0.060 & 0.118 & 0.20\% \\
        \bottomrule
    \end{tabular}
\end{table}

\subsection{LiDAR Targets}
\label{app:data:lidar}

\noindent\textbf{Projection.}
For every frame of a window and every view, the frame's LiDAR sweep is projected into the camera through the vehicle's own model.
Returns on the ego vehicle are removed, the remainder are transformed into the camera frame and restricted to optical-axis depths in $[0.3,80]$\,m, and points outside the lens field of view are discarded before distortion is applied.
Each surviving return is assigned to the pixel containing its projection, and where several returns land on one pixel the nearest is kept.

\noindent\textbf{Pooling.}
The projected returns are pooled onto the latent grid of Sec.~\ref{app:method:architecture}, whose $16\times16$-pixel cells coincide with the cells of the VAE.
For each cell we store the number of returns and the mean of their log depths; the CPP target $d_t^{L,v}(u)$ of Eq.~(\ref{eq:method:cpp_geometry}) is the exponential of the latter, the geometric-mean depth, and a cell is a valid correspondence when it holds at least $n_{\min}=3$ returns.
Cells on moving objects are retained: at each anchor, generated and measured geometry describe the same instant, so the comparison is valid for dynamic and static structure alike.
The same pooled targets serve to fit the depth head of Sec.~\ref{app:method:cpp}.

\noindent\textbf{Poses and rays.}
The recorded poses $\mathbf T^\star_t$ are the front-camera poses of Sec.~\ref{app:data:windows} expressed in $\mathcal C_0$, and equal the composition of the recorded motion rows.
The calibrated rays of Sec.~\ref{app:method:architecture} are obtained by inverting the vehicle's distortion model at every cell and patch centre.

\subsection{Obstacle and Drivable-Area Labels}
\label{app:data:hinges}

\noindent\textbf{Obstacle sets.}
For each future frame $t=1,\ldots,T$, the box set $\mathcal O_t$ of Sec.~\ref{app:method:hinges} contains all annotated objects within 40\,m of the ego vehicle at $t$, of every class, as oriented rectangles in the ground plane of $\mathcal C_0$.
The static set $\mathcal S_t$ is built from the sweep recorded at $t$: returns on the ego vehicle and inside any annotated box, dilated by 0.3\,m, are removed; the remainder are restricted to heights between 0.3\,m and 3\,m above the local ground, estimated from the lowest returns near the vehicle, and to planar distances between 1\,m and 25\,m from the ego position at $t$.
Isolated returns are suppressed by keeping only $0.5$\,m ground-plane columns with at least two returns, each represented by its return nearest to the recorded waypoint, and the set is limited to the 2{,}048 points nearest to that waypoint, so the return that determines the recorded clearance is always among them.
The recorded clearance $d^\star_t$ is evaluated from these sets at the recorded waypoint.

\noindent\textbf{Drivable-area field.}
The drivable region $\mathcal D$ is the union of the road-block, intersection, and car-park polygons of the nuPlan map within 100\,m of the ego vehicle at the current frame, the same layers that the NAVSIM drivable-area metric treats as permissible.
The polygons are rasterized in the ground plane of $\mathcal C_0$ on the $128\times128$ grid of Sec.~\ref{app:method:hinges} with $r_S=0.5$\,m, covering $x\in[-32,32)$\,m and $z\in[-8,56)$\,m, and the signed distance field is computed by Euclidean distance transforms of the rasterized region and its complement.
On 60 held-out windows the rasterized region agrees with the NAVSIM devkit's own drivable map cell for cell.

\noindent\textbf{Coverage.}
Every training window carries all three cameras on all frames, a LiDAR sweep on the current and future frames, calibration, annotations, and a complete set of labels; windows missing any of these are excluded rather than substituted.

%% file: sections/appendix_experiments.tex
\section{Experimental Details and Additional Results}
\label{app:experiments}

\subsection{Hyperparameters and Metrics}
\label{app:experiments:hparams}

\method{} is obtained by full fine-tuning of the generation pathway of Cosmos~3 Nano on the NAVSIM \texttt{navtrain} windows of App.~\ref{app:data}, with the understanding pathway, the VAE, and the depth head fixed.
Every update minimizes the objective of Eq.~(\ref{eq:method:total_objective}): flow matching on the three RGB streams, the three depth streams, and the ego-motion rows, with the CPP and hinge terms scaled to a fixed fraction of the action flow-gradient norm.
The hinge fractions are annealed to zero during the first half of training and CPP stays active throughout, so the geometric supervision that remains at the end is the one the model carries into planning.
One model, evaluated with an exponential moving average of its weights, produces every \method{} result of Sec.~\ref{sec:exp:planning}, and Table~\ref{tab:app:hparams} lists its settings. The rows marked ``without CPP'' come from the same recipe with the CPP term removed (Sec.~\ref{sec:exp:ablations}).

\noindent\textbf{Metrics.}
PDMS combines no collision (NC), drivable-area compliance (DAC), time-to-collision (TTC), comfort (C), and ego progress (EP); EPDMS adds driving-direction (DDC), traffic-light (TLC), and lane-keeping (LK) compliance and splits comfort into history and extended comfort (HC, EC).

\begin{table}[H]
\centering
\caption{\textbf{Hyperparameters} of \method{}.}
\label{tab:app:hparams}
\small
\setlength{\tabcolsep}{8pt}
\renewcommand{\arraystretch}{1.05}
\begin{tabular}{@{}ll@{}}
\toprule
\textbf{Data} & \\
Training windows & 103{,}281 (NAVSIM \texttt{navtrain}) \\
Frames per window & 1 current + 8 future, 2\,Hz \\
Views and resolution & front $832\times468$; front-left and front-right $416\times234$ \\
\midrule
\textbf{Model} & \\
Initialization & Cosmos~3 Nano (15.2B parameters) \\
Trained parameters & generation pathway, 7.0B \\
Noise-level shift & $\sigma=5u/(1+4u)$, $u$ uniform; training and sampling \\
\midrule
\textbf{Optimization} & \\
Updates $\times$ batch & 30{,}000 $\times$ 44 windows \\
Optimizer & AdamW, $\beta=(0.9,\,0.99)$, no weight decay, clip 1.0 \\
Learning rate & peak $2\times10^{-5}$, 300 warm-up updates, linear decay to 0 \\
Precision & fp32 master weights, bf16 compute \\
Weight averaging & EMA, power-function profile, relative width 0.1 \\
\midrule
\textbf{Objective} & \\
$\lambda_V$, $\lambda_A$ & 10, 20 \\
View--modality weights $\omega_{v,m}$ & front RGB 1; side RGB $1/4$; depth $1/6$ per view \\
$\eta_{\mathrm{cpp}}$ & 0.2 throughout \\
$\eta_{\mathrm{obs}}$, $\eta_{\mathrm{drv}}$ & 0.2 to update 9{,}000, annealed smoothly to 0 by update 15{,}000 \\
CPP anchors $\mathcal A$; Huber transition $\delta$ & $t\in\{4,8\}$ (2\,s, 4\,s); 0.5\,m \\
Hinge margins $m_{\mathrm{obs}}$, $m_{\mathrm{drv}}$ & 1.6\,m, 1.5\,m \\
\midrule
\textbf{Inference} & \\
Sampler & UniPC, 30 steps, no guidance \\
Trajectory selection & one sample; or medoid of $K{=}8$ samples \\
\bottomrule
\end{tabular}
\end{table}

\subsection{\texttt{navhard} Stage Scores and Sub-Scores}
\label{app:experiments:navhard}

The two stages of \texttt{navhard} separate the quality of a plan in the recorded scene from its behaviour once traffic reacts to it.
Table~\ref{tab:app:navhard_stages} lists both for the methods that report them.
In the first stage \method{} lies within the range of the other methods; in the second it has the highest no-collision, drivable-area, and time-to-collision terms and the highest stage score.
Each synthetic second-stage scene is weighted by how close its start lies to the end of the first-stage plan, and the combined score multiplies the two stages per scene before averaging, so it is not a function of the two stage means.

\begin{table}[!ht]
\centering
\caption{\textbf{\texttt{navhard} per-stage sub-scores}, as reported; the LTF and DriveVLA-W0 (Flow checkpoint) rows are the values reported in \cite{eponav2}, which reprints the LTF score of \cite{pseudosim}; the DVGT-2 row is as reported in \cite{geowam}. Stage 1: 450 recorded scenes. Stage 2: 5{,}462 synthetic scenes with reactive traffic. Bold: best per column.}
\label{tab:app:navhard_stages}
\footnotesize
\setlength{\tabcolsep}{3pt}
\renewcommand{\arraystretch}{1.0}

\textbf{Stage 1}\\[2pt]
\begin{tabular*}{\linewidth}{@{\extracolsep{\fill}}l ccccccccc c@{}}
\toprule
Method & NC & DAC & DDC & TLC & EP & TTC & LK & HC & EC & Stage 1 \\
\midrule
LTF~\cite{transfuser_pami} & 96.2 & 79.6 & 99.1 & 99.6 & 84.1 & 95.1 & 94.2 & 97.6 & 79.1 & -- \\
DriveVLA-W0~\cite{drivevla_w0} & 96.8 & 83.3 & 99.0 & 99.6 & 84.6 & 95.3 & 96.4 & 97.6 & 78.2 & -- \\
DVGT-2~\cite{dvgt2} & 97.2 & 91.3 & 98.4 & 99.8 & 84.8 & 95.5 & 95.5 & 97.5 & 71.4 & -- \\
DriveFuture~\cite{drivefuture} & 96.3 & 87.8 & 98.2 & 99.6 & 83.1 & 96.9 & 94.9 & 97.6 & 76.9 & -- \\
4D-WAM~\cite{fourdwam} & 92.8 & 89.6 & \textbf{99.7} & 99.3 & \textbf{98.6} & 91.6 & 86.9 & 97.1 & \textbf{81.3} & -- \\
EponaV2~\cite{eponav2} & 97.3 & 90.7 & 99.4 & \textbf{100} & 83.3 & \textbf{97.3} & \textbf{97.3} & 97.6 & 60.9 & -- \\
GeoWAM~\cite{geowam} & \textbf{97.7} & \textbf{91.5} & 99.1 & 99.8 & 83.8 & 95.8 & 96.0 & \textbf{97.8} & 79.0 & -- \\
\midrule
\textbf{\method{}} & 96.9 & 90.9 & 99.0 & 99.6 & 83.8 & 95.6 & 95.8 & \textbf{97.8} & 67.6 & 77.0 \\
\method{}, medoid-of-8 & 96.4 & 91.1 & 99.3 & \textbf{100.0} & 83.6 & 96.4 & 96.0 & \textbf{97.8} & 72.4 & \textbf{78.3} \\
\method{}, without CPP & 97.0 & 90.2 & 99.0 & 99.8 & 83.9 & 95.6 & 95.1 & \textbf{97.8} & 67.6 & 76.6 \\
\bottomrule
\end{tabular*}

\medskip
\textbf{Stage 2}\\[2pt]
\begin{tabular*}{\linewidth}{@{\extracolsep{\fill}}l ccccccccc c@{}}
\toprule
Method & NC & DAC & DDC & TLC & EP & TTC & LK & HC & EC & Stage 2 \\
\midrule
LTF~\cite{transfuser_pami} & 77.8 & 70.2 & 84.3 & 98.1 & 85.1 & 75.7 & 45.4 & 95.7 & \textbf{76.0} & -- \\
DriveVLA-W0~\cite{drivevla_w0} & 76.8 & 64.3 & 79.9 & 98.3 & 89.2 & 75.0 & 46.8 & 95.8 & 53.1 & -- \\
DVGT-2~\cite{dvgt2} & 77.8 & 73.8 & 81.3 & 98.3 & 91.5 & 73.2 & 48.0 & 83.9 & 45.1 & -- \\
DriveFuture~\cite{drivefuture} & 82.3 & 78.8 & \textbf{88.1} & 98.4 & 83.6 & 79.6 & 47.6 & \textbf{97.0} & 75.9 & -- \\
4D-WAM~\cite{fourdwam} & 82.6 & 71.8 & 86.9 & 98.2 & \textbf{97.6} & 78.8 & 49.0 & 95.8 & 71.6 & -- \\
EponaV2~\cite{eponav2} & 83.6 & 78.0 & 88.0 & \textbf{98.9} & 86.0 & 80.3 & 50.1 & 96.1 & 52.0 & -- \\
GeoWAM~\cite{geowam} & 80.4 & 76.3 & 87.3 & 98.7 & 88.9 & 76.2 & 49.9 & 94.0 & 56.0 & -- \\
\midrule
\textbf{\method{}} & 83.9 & 80.2 & 87.3 & 98.7 & 88.7 & 81.6 & 48.9 & 95.3 & 54.4 & 48.8 \\
\method{}, medoid-of-8 & \textbf{85.0} & \textbf{81.7} & \textbf{88.1} & 98.7 & 88.0 & \textbf{82.2} & \textbf{50.5} & 95.5 & 58.4 & \textbf{50.3} \\
\method{}, without CPP & 83.7 & 79.6 & 86.1 & 98.7 & 89.1 & 80.7 & 48.5 & 95.0 & 53.5 & 47.0 \\
\bottomrule
\end{tabular*}
\end{table}

\subsection{HUGSIM by Difficulty, Metric, and Source Dataset}
\label{app:experiments:hugsim}
The simulator renders each source dataset's own camera rig; the model receives the render's pinhole rays, the simulator's navigation command and a 0.5\,s odometry history, and returns an eight-waypoint plan every 0. The simulator advances only after the planner returns, so the 0.5\,s replanning interval is simulation time; at 9.4 GPU-seconds per plan (App.~\ref{app:experiments:selection}) the evaluation does not establish real-time operation.5\,s, which the benchmark's controller tracks. The rendered rig differs from the training rig in lens model, mounting height and appearance, and no adapter is used.

Table~\ref{tab:app:hugsim_tiers} gives every HUGSIM metric by difficulty for \method{} and for the baselines whose per-metric values the benchmark reports.
\method{} has the highest drivable-area compliance at every difficulty and, on easy scenarios, leads every metric except comfort; on medium, hard, and extreme scenarios UniAD keeps higher no-collision and time-to-collision terms, which is where \method{}'s score is lost.
Table~\ref{tab:app:hugsim_datasets} splits \method{} by the source dataset of the reconstructed scenes: route completion and HD-Score are highest on nuScenes and Waymo and lowest on the hard and extreme KITTI-360 scenarios.
HD-Score multiplies route completion by the per-step product of the no-collision and drivable-area terms with a weighted average of the time-to-collision and comfort terms, as PDMS does.

\begin{table}[!ht]
\centering
\caption{\textbf{HUGSIM by difficulty and metric} ($\times 100$); baselines as reported by the benchmark paper. Overall follows the benchmark's reduction: per-dataset episode means, averaged with equal weight over the four source datasets within each difficulty, then weighted over the difficulties by their episode counts (80, 157, 96, 103). Bold: best per metric and difficulty among UniAD, VAD, LTF, and \method{}.}
\label{tab:app:hugsim_tiers}
\footnotesize
\setlength{\tabcolsep}{5pt}
\renewcommand{\arraystretch}{1.0}
\begin{tabular*}{\linewidth}{@{\extracolsep{\fill}}l l cccc cc@{}}
\toprule
Method & Difficulty & NC & DAC & TTC & COM & RC & HD-Score \\
\midrule
UniAD~\cite{uniad} & Easy    & 77.4 & 88.5 & 70.8 & 82.8 & 58.6 & 48.7 \\
                   & Medium  & \textbf{72.4} & 86.8 & \textbf{60.4} & 72.0 & 41.2 & 29.5 \\
                   & Hard    & \textbf{66.5} & 86.0 & \textbf{55.0} & 67.2 & \textbf{40.4} & \textbf{27.3} \\
                   & Extreme & \textbf{54.4} & 89.6 & \textbf{42.9} & 58.5 & 26.0 & \textbf{14.3} \\
\midrule
VAD~\cite{vad}     & Easy    & 66.1 & 73.9 & 58.2 & \textbf{100.0} & 38.7 & 24.3 \\
                   & Medium  & 45.7 & 79.8 & 29.0 & \textbf{100.0} & 27.0 & 9.9 \\
                   & Hard    & 44.3 & 81.0 & 28.8 & \textbf{100.0} & 25.5 & 10.4 \\
                   & Extreme & 36.1 & 83.7 & 25.9 & \textbf{100.0} & 23.0 & 8.2 \\
\midrule
LTF~\cite{transfuser_pami} & Easy    & 74.1 & 81.7 & 70.6 & 99.6 & 68.4 & 52.8 \\
                   & Medium  & 47.4 & 83.2 & 43.3 & 99.7 & 40.7 & 24.6 \\
                   & Hard    & 40.7 & 83.9 & 37.0 & 99.9 & 36.9 & 19.8 \\
                   & Extreme & 28.3 & 85.1 & 21.8 & 99.8 & 25.5 & 8.1 \\
\midrule
\textbf{\method{}} & Easy    & \textbf{93.1} & \textbf{99.1} & \textbf{91.7} & 99.4 & \textbf{93.5} & \textbf{86.9} \\
                   & Medium  & 51.8 & \textbf{97.5} & 39.9 & 95.6 & \textbf{47.4} & \textbf{30.1} \\
                   & Hard    & 47.7 & \textbf{96.0} & 39.1 & 95.8 & 38.6 & 25.2 \\
                   & Extreme & 34.3 & \textbf{93.8} & 27.8 & 95.5 & \textbf{26.1} & 13.4 \\
                   & Overall & 54.4 & 96.6 & 46.4 & 96.3  & 48.9 & 35.5 \\
\midrule
\method{}, without CPP & Easy & 92.7 & 98.8 & 91.0 & 99.3 & 92.6 & 85.4 \\
                   & Medium  & 49.3 & 97.4 & 37.3 & 95.6 & 44.9 & 26.8 \\
                   & Hard    & 47.0 & 96.0 & 37.8 & 95.2 & 37.8 & 23.9 \\
                   & Extreme & 32.6 & 93.9 & 25.7 & 95.6 & 25.4 & 12.0 \\
                   & Overall & 52.8 & 96.5 & 44.5 & 96.2 & 47.5 & 33.4 \\
\bottomrule
\end{tabular*}
\end{table}

\begin{table}[!ht]
\centering
\caption{\textbf{HUGSIM by source dataset}, \method{}: RC / HD-Score per difficulty level and the plain episode mean per dataset. The Overall values of Table~\ref{tab:app:hugsim_tiers} average these per-dataset means with equal dataset weight within each difficulty before weighting by episodes, so they are not the pooled episode mean.}
\label{tab:app:hugsim_datasets}
\footnotesize
\setlength{\tabcolsep}{5pt}
\renewcommand{\arraystretch}{1.0}
\begin{tabular*}{\linewidth}{@{\extracolsep{\fill}}l c cccc c@{}}
\toprule
Dataset & Episodes & Easy & Medium & Hard & Extreme & All \\
\midrule
nuScenes  & 88  & 94.5 / 88.4 & 43.7 / 27.1 & 59.5 / 43.1 & 40.6 / 27.3 & 56.7 / 42.9 \\
Waymo     & 108 & 94.6 / 91.6 & 51.5 / 34.1 & 57.3 / 44.8 & 23.9 / 8.7  & 55.8 / 42.8 \\
KITTI-360 & 113 & 87.4 / 76.5 & 51.6 / 39.3 & 14.9 / 6.1  & 8.0 / 0.7   & 41.1 / 31.0 \\
PandaSet  & 127 & 97.4 / 91.1 & 42.8 / 20.1 & 22.5 / 6.6  & 32.0 / 16.9 & 41.3 / 24.3 \\
\bottomrule
\end{tabular*}
\end{table}

\subsection{Ablation Details}
\label{app:experiments:ablations}

\noindent\textbf{LoRA setting.}
Table~\ref{tab:app:hparams_lora} lists the setting of the ablation rows in Table~\ref{tab:lora}b and~d: rank-128 adapters ($\alpha=256$) on the attention projections of the generation pathway, trained for 16{,}000 updates.
Settings not listed are as in Table~\ref{tab:app:hparams}.
The reference row is flow + hinge + CPP with all three views, and every other row changes only the objective terms or inputs given in its label.
Removing generated depth also removes CPP, which acts on it.
Table~\ref{tab:app:lora_full} gives all sub-scores and Table~\ref{tab:app:lora_cmd} the scores by driving command.
Without generated depth, and hence without CPP, the multi-view model scores 87.5 PDMS, 0.3 above flow + hinge with generated depth and 0.7 below the full objective.
Generated depth therefore helps planning through CPP rather than on its own.

\begin{table}[!htb]
\centering
\caption{\textbf{Hyperparameters of the LoRA setting.} Adapters act on the generation pathway; rows not listed are as in Table~\ref{tab:app:hparams}.}
\label{tab:app:hparams_lora}
\small
\setlength{\tabcolsep}{8pt}
\renewcommand{\arraystretch}{1.05}
\begin{tabular}{@{}ll@{}}
\toprule
Adaptation & LoRA, rank 128, $\alpha=256$, on the attention projections \\
Trained parameters & 122.7M \\
Updates $\times$ batch & 16{,}000 $\times$ 44 windows \\
Optimizer & AdamW, $\beta=(0.9,\,0.99)$, weight decay 0.05, clip 1.0 \\
Learning rate & peak $2\times10^{-4}$, 300 warm-up updates, linear decay to 0 \\
Weight averaging & none \\
\bottomrule
\end{tabular}
\end{table}

\begin{table}[!ht]
\centering
\caption{\textbf{LoRA setting, all sub-scores} on \texttt{navtest}, one sample per scene.}
\label{tab:app:lora_full}
\footnotesize
\setlength{\tabcolsep}{2pt}
\renewcommand{\arraystretch}{1.0}
\begin{tabular*}{\linewidth}{@{\extracolsep{\fill}}l cccccccc cc@{}}
\toprule
Row & NC & DAC & DDC & TLC & EP & TTC & LK & HC & PDMS & EPDMS \\
\midrule
flow                              & 95.9 & 96.4 & 98.6 & 99.7 & 86.0 & 94.9 & 95.2 & 98.4 & 85.7 & 84.3 \\
flow + hinge                      & 96.8 & 96.6 & 98.5 & 99.8 & 86.8 & 95.9 & 95.3 & 98.6 & 87.2 & 85.7 \\
flow + CPP                        & 96.8 & 97.0 & 98.5 & 99.7 & 87.0 & 95.7 & 95.1 & 98.6 & 87.8 & 86.2 \\
flow + hinge + CPP (multi-view)   & 96.9 & 97.2 & 98.5 & 99.7 & 87.2 & 95.8 & 95.1 & 98.6 & \textbf{88.2} & \textbf{86.5} \\
\midrule
single-view                       & 96.6 & 96.7 & 98.6 & 99.7 & 86.8 & 95.6 & 95.2 & 98.5 & 87.1 & 85.6 \\
multi-view, no depth              & 96.9 & 96.8 & 98.5 & 99.7 & 87.0 & 95.9 & 95.2 & 98.6 & 87.5 & 85.9 \\
\bottomrule
\end{tabular*}
\end{table}

\begin{table}[!ht]
\centering
\caption{\textbf{Scores by driving command} on \texttt{navtest} (2{,}501 left, 8{,}070 straight, 1{,}575 right scenes), one sample per scene.}
\label{tab:app:lora_cmd}
\footnotesize
\setlength{\tabcolsep}{5pt}
\renewcommand{\arraystretch}{1.0}
\begin{tabular*}{\linewidth}{@{\extracolsep{\fill}}l ccc ccc@{}}
\toprule
 & \multicolumn{3}{c}{PDMS} & \multicolumn{3}{c}{EPDMS} \\
\cmidrule(lr){2-4}\cmidrule(l){5-7}
Row & Left & Straight & Right & Left & Straight & Right \\
\midrule
flow                              & 81.4 & 87.8 & 81.9 & 80.3 & 86.2 & 81.3 \\
flow + hinge                      & 83.4 & 89.0 & 83.8 & 82.4 & 87.2 & 82.8 \\
flow + CPP                        & 83.5 & 89.9 & 83.9 & 82.1 & 88.1 & 83.0 \\
flow + hinge + CPP (multi-view)   & 83.9 & 90.3 & 84.3 & 82.4 & 88.4 & 83.3 \\
\midrule
single-view                       & 81.2 & 90.0 & 82.0 & 80.1 & 88.2 & 81.4 \\
multi-view, no depth              & 83.5 & 89.4 & 84.0 & 82.4 & 87.6 & 83.0 \\
\midrule
\method{} (full fine-tune)        & 87.8 & 92.7 & 90.4 & 87.3 & 91.4 & 89.4 \\
\bottomrule
\end{tabular*}
\end{table}

\begin{table}[!htb]
\centering
\caption{\textbf{Generated depth with and without CPP} for the final models, on \texttt{navtest}: AbsRel and $\delta_{1.25}$ against the LiDAR of the same future frame and camera, without scale alignment, as in Table~\ref{tab:worldmodel}a; means over four samples per scene. ``Cells'': the 16-pixel cells that CPP supervises.}
\label{tab:app:cpp_depth}
\footnotesize
\setlength{\tabcolsep}{4pt}
\begin{tabular}{@{}l cc cc cc cc@{}}
\toprule
 & \multicolumn{4}{c}{+2\,s} & \multicolumn{4}{c}{+4\,s} \\
\cmidrule(lr){2-5}\cmidrule(l){6-9}
 & \multicolumn{2}{c}{AbsRel$\downarrow$} & \multicolumn{2}{c}{$\delta_{1.25}\uparrow$} & \multicolumn{2}{c}{AbsRel$\downarrow$} & \multicolumn{2}{c}{$\delta_{1.25}\uparrow$} \\
\cmidrule(lr){2-3}\cmidrule(lr){4-5}\cmidrule(lr){6-7}\cmidrule(l){8-9}
View & with & without & with & without & with & without & with & without \\
\midrule
Front        & 0.175 & 0.193 & 0.814 & 0.802 & 0.232 & 0.253 & 0.742 & 0.743 \\
Front-left   & 0.292 & 0.315 & 0.725 & 0.706 & 0.374 & 0.404 & 0.654 & 0.649 \\
Front-right  & 0.240 & 0.277 & 0.752 & 0.725 & 0.305 & 0.346 & 0.676 & 0.663 \\
Front, cells & 0.144 & 0.154 & 0.856 & 0.844 & 0.194 & 0.205 & 0.795 & 0.795 \\
\bottomrule
\end{tabular}
\end{table}

\noindent\textbf{Model trained without CPP.}
At 4{,}000 updates (Table~\ref{tab:cpp}a, top), the plans of the model without CPP drift laterally through a less accurate heading rather than taking wrong turns.
The final model without CPP also generates worse depth (Table~\ref{tab:app:cpp_depth}).
Its AbsRel against LiDAR is higher on every view and horizon, by 8--15\% at the pixel level and 5--7\% on the cells that CPP supervises, and its $\delta_{1.25}$ is lower on every view and horizon except the front view at +4\,s, where the two are equal within sample noise.
In the front view at +2\,s, AbsRel is 0.193 against 0.175.
Of its sampled futures, 70.9\% end at +4\,s within $10^\circ$ of the recorded heading and within 0.3\,m of the recorded position along the forward axis, against 81.4\% with CPP.
Video quality (FVD, FID) is unchanged within sample noise.
Removing CPP from update 20{,}000 onward leaves open-loop scores within sample noise but costs 1.2 HD-Score and 1.1 RC on HUGSIM, and raises the share of wrong-turn futures, those with heading error above $40^\circ$ at +4\,s, from 0.4\% to 0.6\%.

\noindent\textbf{Hinge schedule.}
Both hinges are one-sided and vanish on every plan that keeps its margins, and the obstacle margin is capped at the recorded clearance (Sec.~\ref{sec:method:hinges}), so the recorded trajectory incurs no obstacle loss.
Early in training many plans violate the margins, and the hinges give a direct gradient on waypoint positions.
Late in training few plans do, yet the balancing of Sec.~\ref{sec:method:training_inference} still scales each active term to a fixed fraction of the action flow-gradient norm, which would concentrate that fraction on the few remaining plans.
We therefore anneal both hinges to zero by the middle of training.
CPP penalizes deviation in either direction on every LiDAR-supported cell, so it does not vanish once plans are safe.

\subsection{World-Model Quality and Consistency}
\label{app:experiments:consistency}

\begin{table}[!htb]
\centering
\caption{\textbf{Front-view video quality on NAVSIM.} I3D FVD over one context and eight generated frames at 2\,Hz, FID over the same frames. 600 clips: generated clips of half the scenes against the recorded clips of the other half, ten splits, floor from the two recorded halves; 1{,}200 clips: generated against recorded clips of the same scenes. Reported rows use their own clip counts and lengths (DrivingGPT~\cite{drivinggpt} 12 frames; PWM~\cite{pwm} 10, $\dagger$ as reported by DriveDreamer-Policy~\cite{drivedreamer_policy}, itself 9; CoWorld-VLA~\cite{coworldvla} unstated).}
\label{tab:app:video}
\footnotesize
\setlength{\tabcolsep}{5pt}
\begin{tabular}{@{}l c cc@{}}
\toprule
 & Clips & FVD$\downarrow$ & FID$\downarrow$ \\
\midrule
DrivingGPT & 512 & 142.6 & 12.8 \\
PWM$^\dagger$ & -- & 86.0 & -- \\
DriveDreamer-Policy & -- & 53.6 & -- \\
CoWorld-VLA & -- & 32.7 & -- \\
\midrule
Recorded (floor) & 600 & 91.5{\tiny$\pm$3.2} & 11.0{\tiny$\pm$0.3} \\
\method{} & 600 & 111.3{\tiny$\pm$6.8} & 15.4{\tiny$\pm$0.6} \\
\method{} & 1{,}200 & 42.2 & 6.8 \\
\bottomrule
\end{tabular}
\end{table}

\noindent\textbf{Metric depth.}
Decoded from a sample and scored against the LiDAR sweep of the same future frame and camera without any scale alignment, front-view depth reaches AbsRel 0.175 at +2\,s and 0.232 at +4\,s, with $\delta_{1.25}$ of 0.814 and 0.742 (Table~\ref{tab:worldmodel}c); on the cells that CPP supervises, AbsRel is 0.144 and 0.194.
The side views, generated at half resolution, have 1.3--1.7$\times$ the front error (Table~\ref{tab:app:cpp_depth}), and the median signed error is negative in every front-view setting, placing generated surfaces 0.2--0.3\,m nearer than the LiDAR.
Because each sample renders the model's own future viewpoint, these errors include any difference between the generated and the recorded ego motion.
Reported future-depth results on nuScenes (Table~\ref{tab:worldmodel}c, lower block) place GeoWAM~\cite{geowam} at AbsRel 0.245 and 0.297 at the two horizons, Epona~\cite{epona} with a DVGT~\cite{dvgt} geometry head at 0.263 and 0.310, VGGT-World~\cite{vggtworld} at 0.329 and 0.357, and the same base model as ours with that head, Cosmos~3~\cite{cosmos3} + DVGT, at 0.376 and 0.422; the front-view values of \method{} are lower at both horizons, on a different dataset and rig.

\noindent\textbf{Video.}
Table~\ref{tab:app:video} gives the front-view video scores behind Sec.~\ref{sec:exp:world}.
At 600 clips per side, the generated front video reaches FVD-9 111.3 against a recorded-versus-recorded floor of 91.5, an excess of 19.8 that is positive on all ten splits, and FID 15.4 against 11.0; over all 1{,}200 clips it reaches 42.2 and 6.8 (Table~\ref{tab:app:video}).
Reported NAVSIM front-view results range from FVD 142.6 for DrivingGPT~\cite{drivinggpt} on 512 clips to 32.7 for CoWorld-VLA~\cite{coworldvla}, each under its own clip count and length; on nuPlan, the source data of NAVSIM, Epona~\cite{epona} reports 50.8 over 1{,}628 ten-frame clips.
At a similar clip count, \method{} is below DrivingGPT, and the recorded floor shows how much of any reported value is sample size: the gap between generated and recorded clips is about a fifth of the floor itself.

\begin{table}[!htb]
\centering
\caption{\textbf{Consistency of the generated future} with its own motion, across views, and over time, on \texttt{navtest}; medians. Yaw: camera rotation recovered from the generated video~\cite{mapanything} against the generated motion, and, for the floor, from the recorded clip against the recorded trajectory; the excess is the median of the per-scene paired differences, with a 95\% bootstrap interval. Depth: median $|\Delta\log z|$ after warping the front depth into the side views with the recorded extrinsics, or frame 0 into frame 8 with the generated motion; floors from two LiDAR sweeps 0.5\,s apart and from recorded LiDAR under recorded motion.}
\label{tab:app:consistency}
\footnotesize
\setlength{\tabcolsep}{4pt}
\begin{tabular}{@{}l c c c@{}}
\toprule
Quantity & Floor & \method{} & Excess \\
\midrule
Yaw, video vs.\ motion, 4\,s ($^\circ$) & 0.29 & 0.80 & +0.44 [+0.39, +0.48] \\
\quad turns above $45^\circ$ & 6.11 & -- & +2.31 [+1.93, +2.86] \\
Cross-view $|\Delta\log z|$, +2\,s, L\,/\,R & 0.032\,/\,0.030 & 0.045\,/\,0.049 & 1.5$\times$ \\
Temporal $|\Delta\log z|$, 0 to 4\,s & 0.087 & 0.285 & 3.3$\times$ \\
\bottomrule
\end{tabular}
\end{table}

\noindent\textbf{Consistency with the generated motion.}
Table~\ref{tab:app:consistency} gives the floors and excesses behind the consistency line of Sec.~\ref{sec:exp:world}.
The camera yaw that MapAnything~\cite{mapanything} recovers from each generated video deviates from the generated motion by a median of $0.80^\circ$ over 4\,s, against $0.29^\circ$ when it is run on recorded clips, and where the generated motion departs from the recorded trajectory the video follows the generated motion.
The generated front depth agrees with the generated side depth to within 1.5$\times$ the disagreement between two LiDAR sweeps 0.5\,s apart, and frame to frame under the generated motion the disagreement grows to 3.3$\times$ the LiDAR floor over 4\,s.
The yaw excess over the 4\,s horizon is $0.44^\circ$ with a 95\% bootstrap interval of $[0.39, 0.48]$; above $45^\circ$, where the median commanded turn is $58.7^\circ$, it is $2.31^\circ$.
The excess concentrates in sharp turns, where the video realizes about 96\% of a hard turn; where the generated motion and the recorded trajectory differ by more than $2^\circ$, the video follows the generated motion in 67\% of the scenes.
Warped into the side views with the recorded extrinsics, the generated front depth disagrees with the generated side depth by a median $|\Delta\log z|$ of 0.045--0.049 at +2\,s, about 1.5$\times$ the disagreement between two LiDAR sweeps 0.5\,s apart.
Warped from frame to frame with the generated motion, the per-step depth disagreement stays between 0.038 and 0.043 and accumulates close to linearly to 0.285 over 4\,s, 2.3--3.3$\times$ the LiDAR floor along the horizon.
At every step the generated depth agrees more closely with the generated motion than with the recorded motion, by about 20\%.

\subsection{Sampler Steps and Cost}
\label{app:experiments:selection}

\begin{table}[t]
\centering
\caption{\textbf{Sampler steps and cost} for the model of Tables~\ref{tab:navsim_v1}--\ref{tab:hugsim}: PDMS relative to the 30-step protocol row, one sample per scene, averaged over the number of samples given (sample-to-sample sd 0.2--0.3 where repeated), and the cost of one plan on one RTX PRO 6000 GPU.}
\label{tab:app:selection}
\footnotesize
\setlength{\tabcolsep}{6pt}
\begin{tabular}{@{}c c c c@{}}
\toprule
UniPC steps & $\Delta$PDMS & Samples & GPU-s / scene \\
\midrule
4  & $-0.6$ & 1 & 3.5 \\
8  & $+0.7$ & 3 & 4.4 \\
15 & $-0.4$ & 1 & 6.0 \\
30 & 0      & 4 & 9.4 \\
\bottomrule
\end{tabular}
\end{table}

Table~\ref{tab:app:selection} varies the sampler steps at inference.
The step count moves PDMS by less than a point and not monotonically: 8 steps score 0.7 above the 30-step protocol and the single 4- and 15-step samples 0.6 and 0.4 below, against a sample-to-sample sd of 0.2--0.3, while the cost of a plan grows from 3.5 to 9.4 GPU-seconds per scene.
All other results in this paper use 30 steps.

\subsection{Additional Qualitative Examples}
\label{app:experiments:qualitative}

\begin{figure}[p]
\centering
\includegraphics[width=\linewidth]{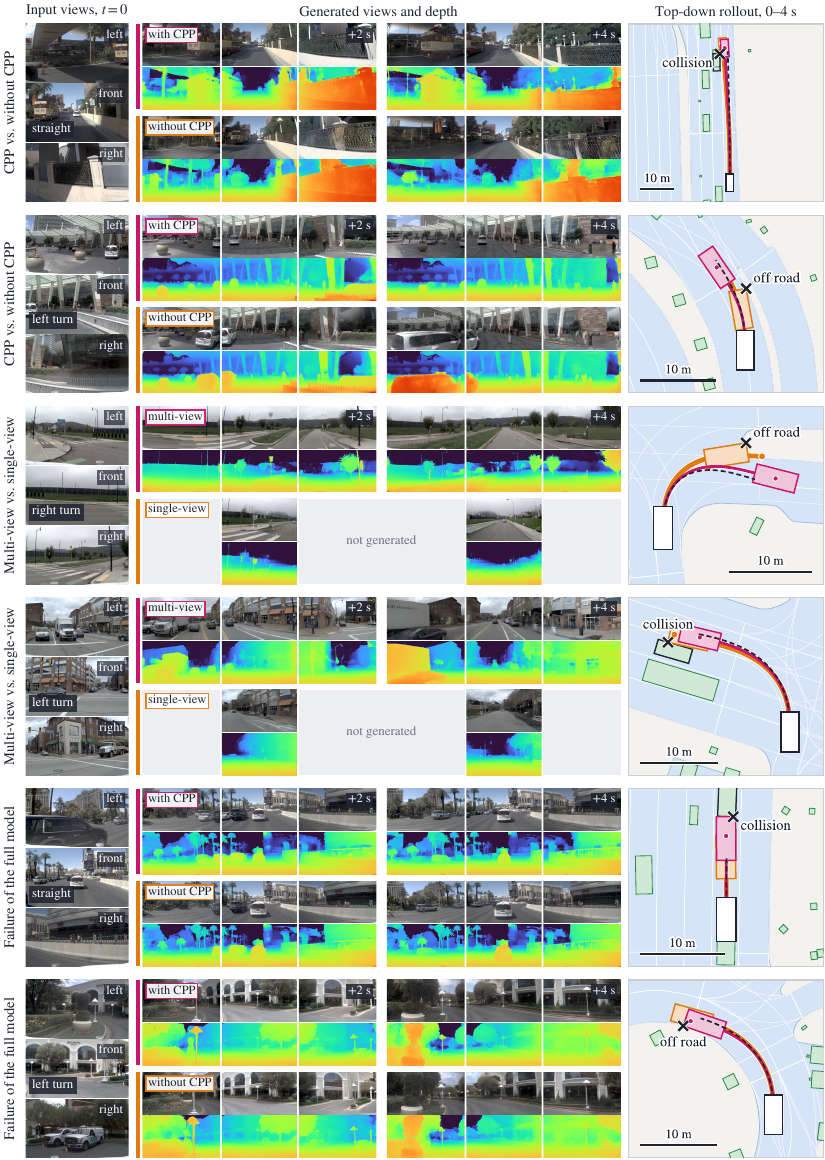}
\caption{\textbf{Additional qualitative examples.} On a straight road, the model with CPP generates the overtaking bus alongside in its left view. The model trained without CPP omits it and is clipped by the overtaking car at 3.9\,s. At a hotel drop-off, it generates the departing minivan too close and clips the island at 1.6\,s. The single-view model swings wide over a grass median that only the right camera shows and leaves the road at 3.1\,s. At a left turn, it never generates the waiting truck and hits it at 3.7\,s. The full model renders a stopped minivan at a constant size, plans as if it were pulling away, and rear-ends it at 3.7\,s, a world-model error. At a tight left turn, both models generate indistinguishable views, but the full model clips the inner curb by 0.4\,m at 3.6\,s, a plan-precision error.}
\label{fig:app:qualitative}
\end{figure}

Figure~\ref{fig:app:qualitative} compares the model with CPP against the model trained without CPP, and the multi-view model against the single-view model, on two further scenes each.
It also shows two scenes in which the full model fails.

%% file: sections/appendix_limitations.tex
\section{Limitations}
\label{app:limitations}

\noindent\textbf{Supervision.}
CPP compares generated geometry with LiDAR sweeps carried by the recorded poses, so training needs a LiDAR sensor calibrated to the cameras and an accurate ego trajectory.
The depth streams need dense metric labels, which we obtain from a reconstruction model anchored to the same LiDAR (App.~\ref{app:data:depth}), and the hinges need obstacle and drivable-area labels (App.~\ref{app:data:hinges}).
Every term of the objective is therefore tied to an instrumented collection platform.
Scaling to the far larger body of driving video recorded without LiDAR would need a self-supervised form of the coupling, for instance between generated depth and generated motion alone, which we have not developed.

\noindent\textbf{Consistency against the recorded future only.}
CPP enforces agreement between generated depth and motion where LiDAR supports the recorded trajectory.
Self-consistency away from that trajectory, under counterfactual commands or beyond the 4\,s window, is measured only indirectly (App.~\ref{app:experiments:consistency}) and is not supervised, and whether the coupling generalizes there is open.

\noindent\textbf{One backbone, one scale, one training run per configuration.}
All results come from Cosmos~3 Nano at one model size, and each training configuration was run once.
We report sample-to-sample variation (Table~\ref{tab:navsim_v1}) but not run-to-run variation, and we have not studied how the formulation behaves across backbones, model sizes or data scales due to compute constraints.

\noindent\textbf{Inference cost.}
Every plan generates a full multi-view future.
One plan costs 9.4 GPU-seconds at 30 sampler steps and 4.4 at 8 (App.~\ref{app:experiments:selection}), the medoid rule multiplies this by the number of samples, and an action-only inference path was not tested; the closed-loop evaluation replans in simulation time and does not establish real-time operation.

%% file: sections/appendix_related_work.tex
\section{Extended Related Work}
\label{app:related_work}

\input{paradigm_comparison}

Table~\ref{tab:paradigms} summarizes how representative works connect future prediction to planning; the paragraphs below discuss them in turn.

\noindent\textbf{Generative driving world models.}
Driving world models learn controllable future-video generation from driving data~\cite{drivedreamer,gaia1,vista}, with GAIA-2 supporting coordinated generation across multiple cameras~\cite{gaia2}.
Other representations make scene structure explicit: MUVO predicts images, LiDAR, and occupancy~\cite{muvo}, while OccWorld jointly forecasts occupancy and ego motion~\cite{occworld}.
OmniNWM further combines panoramic video with depth, semantics, and occupancy~\cite{omninwm}.
These approaches establish the value of modeling the future beyond appearance (RGB video) alone.
\method{} combines multiview video and metric depth with ego-motion generation in a shared flow-matching model, providing explicit scene geometry alongside the generated action.

\noindent\textbf{World-action models (WAMs) and planning.}
Future prediction can improve policy representations through auxiliary visual or latent objectives~\cite{law,simwam}.
Joint video--action modeling has also been explored in driving~\cite{epona,driveva} and robotics~\cite{uwm,dreamzero}.
Beyond generation, predicted visual or latent futures support learned trajectory evaluation and policy refinement~\cite{world4drive,resim,driveworld_vla}.
ExploreVLA uses uncertainty in future predictions to guide safety-aware policy exploration~\cite{explorevla}.
\method{} follows the joint-generation approach while keeping trajectory selection simple: a parameter-free consensus rule uses only pairwise distances among sampled trajectories, without a learned scorer or simulator feedback.

\noindent\textbf{Geometry in world and action generation.}
View synthesis provides a foundation for jointly learning depth and camera motion through geometric supervision~\cite{sfmlearner}.
In generative models, geometry serves as a conditioning signal, an output, or a training target.
GeoDrive and ReCamDriving use geometry rendered along supplied trajectories to guide video generation~\cite{geodrive,recamdriving}.
UniFuture jointly forecasts RGB and depth~\cite{unifuture}, while X-WAM adds depth prediction to video--action modeling~\cite{xwam}.
DriveDreamer-Policy integrates depth, video, and action experts~\cite{drivedreamer_policy}, and GeoWorldAD and GeoWAM use geometric representations to condition planning~\cite{geoworldad,geowam}.
Geometry Forcing aligns video-model features with geometric representations~\cite{geometryforcing}, while 4D-WAM supervises generated video through depth and features recovered from generated and recorded frames~\cite{fourdwam} via a Geometric Foundation Model \cite{vggt, vggt-omega}.
\method{} explicitly includes generated ego motion in its geometric objective.
CPP unprojects generated depth, transforms the resulting geometry using generated motion, and compares it with LiDAR transformed using recorded motion in a shared reference frame.
This shared metric loss promotes physical consistency with the measured scene by supervising both outputs alongside their individual flow-matching objectives.

%% file: paradigm_comparison.tex
\begin{table}[!htb]
    \centering
    \caption{
        \textbf{World modeling and its role in planning.}
        Representative works illustrate how future prediction
        supports policy learning, action generation, or trajectory
        selection. These roles can overlap within a method.
    }
    \label{tab:paradigms}
    \fontsize{8.5}{9.5}\selectfont
    \setlength{\tabcolsep}{4pt}
    \renewcommand{\arraystretch}{1.08}
    \renewcommand{\tabularxcolumn}[1]{>{\raggedright\arraybackslash}m{#1}}

    \begin{tabularx}{\linewidth}{@{}
        >{\centering\arraybackslash}m{0.26\linewidth}
        >{\centering\arraybackslash}m{0.21\linewidth}
        X@{}}

        \toprule
        \textbf{Representative works}
        & \textbf{Predicted world}
        & \multicolumn{1}{c@{}}{\textbf{Connection to action}} \\
        \midrule

        LAW, SimWAM~\cite{law,simwam}
        & Future features or video
        & Future prediction provides training supervision;
          planning does not require explicit future generation. \\

        \addlinespace[4pt]

        DriveWAM, GeoWAM~\cite{drivewam,geowam}
        & Future video or 3D geometry
        & Representations of the predicted future
          condition trajectory generation. \\

        \addlinespace[4pt]

        Epona, DriveVA~\cite{epona,driveva}
        & Future video
        & Video and action generation share
          learned representations. \\

        \addlinespace[4pt]

        4D-WAM~\cite{fourdwam}
        & Future video
        & Joint video--action learning is supervised by matching
          geometry recovered from generated and recorded video. \\

        \addlinespace[4pt]

        WoTE, DA-WAM~\cite{wote,dawam}
        & Future scene features
        & A learned scorer evaluates candidate trajectories
          using their predicted future states. \\

        \midrule

        \textbf{\method{} (ours)}
        & Multiview video and metric depth
        & Video, depth, and ego motion are co-denoised.
          CPP jointly supervises generated depth and motion
          against measured scene geometry.
          A label-free consensus rule selects among trajectories. \\

        \bottomrule
    \end{tabularx}
\end{table}